\documentclass[11pt,a4paper,twocolumn]{article}

\usepackage[margin=1in]{geometry}
\usepackage[T1]{fontenc}
\usepackage[utf8]{inputenc}
\usepackage{lmodern}
\usepackage{microtype}
\usepackage{graphicx}
\usepackage{booktabs}
\usepackage{tabularx}
\usepackage{array}
\usepackage{amsmath,amssymb,bm}
\usepackage[authoryear,round]{natbib}
\usepackage{xcolor}
\usepackage{dblfloatfix}
\usepackage[section]{placeins}
\usepackage{hyperref}
\let\OriginalIncludeGraphics\includegraphics
\renewcommand{\includegraphics}[2][]{%
  \IfFileExists{#2}{%
    \OriginalIncludeGraphics[#1]{#2}%
  }{%
    \fbox{\parbox{0.9\linewidth}{\centering Missing figure: \texttt{\detokenize{#2}}}}%
  }%
}
\hypersetup{
  colorlinks=true,
  linkcolor=blue,
  citecolor=blue,
  urlcolor=blue
}

\newcommand{\keywords}[1]{\vspace{0.8em}\noindent\textbf{Keywords:} #1\par\vspace{1em}}
\begin{document}

\title{Gaussian Process Prior Variational Autoencoder for Endoscopic Videos}

\author{
Ivan De Boi\thanks{Corresponding author: \href{mailto:ivan.deboi@uantwerpen.be}{ivan.deboi@uantwerpen.be}}\textsuperscript{1},
Xinxing Shi\textsuperscript{2},
Xiaoyu Jiang\textsuperscript{2},
Tim J.M. Jaspers\textsuperscript{3}\\
Francisco Caetano\textsuperscript{3},
Mauricio A. \'Alvarez\textsuperscript{2},
Fons van der Sommen\textsuperscript{3},
Sam Van der Jeught\textsuperscript{1}\\[0.6em]
\small \textsuperscript{1}Department of Electromechanics, InViLab, University of Antwerp, Antwerp, Belgium\\
\small \textsuperscript{2}Department of Computer Science, University of Manchester, Manchester, United Kingdom\\
\small \textsuperscript{3}Department of Electrical Engineering, Eindhoven University of Technology,\\\small Eindhoven, the Netherlands
}

\date{}

\maketitle
\begin{abstract}
Endoscopic video analysis is essential for gastrointestinal diagnosis and computer-assisted interventions, but video sequences are routinely degraded by specular reflections, motion artifacts, and missing frames. These transient corruptions can distract clinicians, reduce image interpretability, and disrupt downstream tasks such as 3D reconstruction and navigation. Effective restoration therefore requires methods that exploit temporal continuity rather than treating frames in isolation. We introduce a Gaussian Process Prior Variational Autoencoder (GPVAE) framework for endoscopic video restoration that replaces the standard factorized latent prior with a temporal Gaussian process prior, enabling interpolation of missing frames with uncertainty-aware reconstruction.

The framework combines endoscopy-specific encoders, including a convolutional EndoVAE backbone and pretrained Vision Transformer encoders from GastroNet-5M, with two scalable GP approximations: Hierarchical Prior Approximation (HPA) and Sparse Precision Approximation (SPA). Specular reflections are handled using a DUCKNet-based masking pipeline that excludes corrupted pixels from the reconstruction objective. On the C3VDv2 colonoscopy dataset, the best GPVAE variants reduced image reconstruction RMSE by 21.9\% on average, and by up to 26.1\%, relative to matched VAE baselines. Downstream trajectory RMSE was reduced by 12.7\% on average across classical visual odometry and a pretrained PoseNet, at an average increase of 27.3\% in training time per epoch. Finally, the GP posterior provides per-frame uncertainty estimates that reflect temporal support and offer a confidence signal for restored frames.
\end{abstract}


\section{Introduction}\label{sec:introduction}

Endoscopy is a cornerstone of modern clinical practice, supporting disease diagnosis and minimally invasive intervention across a broad range of procedures. It plays a central role in gastrointestinal (GI) examination \citep{Ren2025Prompt-basedEndoscopy} and laparoscopic and robotic surgery \citep{Fujii2018GazeSurgery}. Beyond its immediate clinical use, the increasing availability of digital endoscopic video has created important opportunities for computational analysis, including lesion detection \citep{Zhang2024FromImaging}, tissue characterization \citep{Ferreira2020CMRCharacterization}, tracking \citep{Ye2016OnlineExaminations}, surgical scene understanding, and three-dimensional reconstruction \citep{Shen2026UniSurg:Understanding}.

Despite its clinical importance, endoscopic imaging remains particularly challenging. The acquisition process is frequently affected by artifacts such as specular reflections, motion blur, defocus, smoke, bubbles, sensor noise, and transient occlusions. These degradations can obscure diagnostically relevant structures, distort local appearance, and reduce the reliability of downstream methods such as 3D reconstruction and camera path estimation \citep{Yu2026EndoscopicSegmentation}. Among them, specular reflections are especially problematic because they induce saturated image regions that do not faithfully represent tissue content, thereby rendering parts of the image unusable.

Moreover, endoscopic videos may suffer from entirely missing frames due to acquisition instability, rapid camera motion, transmission failures, or temporary loss of visibility. Such corruption does not only affect individual images, but also disrupts the temporal continuity on which many subsequent tasks rely. Consequently, restoration in an endoscopic video should not be treated solely as a frame-wise enhancement problem, but rather as a temporally structured inference problem in which both spatial corruption and inter-frame dependence are modelled explicitly \citep{Atasoy2012EndoscopicBiopsy, Li2020LearningSupervision}.

Deep generative models provide a natural framework for addressing these problems because they learn a compact latent representation of the high-dimensional images. In particular, variational autoencoders (VAEs) offer a mechanism for encoding observations into a latent space and decoding them back into the image domain. For restoration tasks, this latent representation can be encouraged to capture the underlying anatomical structure of the scene while suppressing nuisance variation associated with corruption and acquisition artifacts. This makes VAEs attractive for image denoising, inpainting, and artifact removal in medical imaging, including endoscopic applications \citep{Diamantis2022EndoVAE:Autoencoder}.

However, most existing restoration methods based on autoencoding or image-to-image translation treat frames independently and therefore do not exploit the temporal aspect of a video. Consecutive endoscopic frames are correlated through camera motion, gradual viewpoint changes, and persistent anatomical content. Frames that are close in time should therefore correspond to nearby latent representations. Models that ignore this structure discard a major source of information, particularly when observations are noisy, masked, or incomplete. Temporal extensions of VAEs have been investigated in the broader machine learning literature through the use of structured priors over latent variables. Among these, Gaussian process priors variational autoencoders (GPVAE)\citep{Rasmussen2006GaussianLearning, Casale2018GaussianAutoencoders} are particularly appealing because they define smooth stochastic processes over time and naturally support interpolation between observed samples. Imposing a Gaussian process (GP) prior over the latent sequence therefore transforms the latent representation from a collection of independent points into a continuous probabilistic trajectory. 

Nevertheless, temporally structured latent-variable models have not yet been fully developed for endoscopic restoration in conjunction with encoder-decoder architectures tailored to endoscopic imagery. This is a critical limitation, since endoscopic data differ substantially from natural-image and typical machine learning benchmarks. In this work, we propose a temporally structured variational autoencoding framework for endoscopic videos in which a Gaussian process prior is imposed over the latent variables. The model combines endoscopy-specific visual encoding and decoding with a continuous latent temporal process, thereby enabling mappings from observed frames to latent trajectories and from latent trajectories back to restored images. By explicitly exploiting temporal correlations across neighboring frames, the proposed framework can suppress localized artifacts, infer missing pixel content, and reconstruct corrupted or absent frames.

An additional advantage of the proposed formulation is its probabilistic nature. Rather than producing only a deterministic reconstruction, the model yields uncertainty estimates associated with the inferred latent states and generated images. Such uncertainty information is highly relevant in medical image analysis, where reconstruction quality may vary across frames and anatomical regions, and where confidence awareness is essential for safe downstream use. Predictive uncertainty can therefore serve as an intrinsic indicator of reconstruction reliability, for example by highlighting frames or regions that require clinician review or by informing the use of restored data in subsequent processing pipelines.

The main contributions of this work are as follows. First, we adapt the VAE framework to a GPVAE setting for endoscopic video by replacing the independent latent prior with a temporal Gaussian process prior. This requires tailoring the existing GPVAE formulation to endoscopic image reconstruction, including a weighted objective that combines per pixel mean squared error (MSE) term, a Learned Perceptual Image Patch Similarity (LPIPS) term, and a Kullback-Leibler (KL) term. 
Second, we integrate the temporal Gaussian process prior with endoscopy-specific encoder and decoder architectures, allowing the model to exploit domain-specific inductive biases of endoscopic imagery for more accurate and reliable restoration.
Third, we apply the resulting models to restoration tasks common in endoscopy, including reconstruction of missing frames and pixels. Finally, the GP posterior over the latent trajectory provides an uncertainty estimate for each reconstructed frame.

The rest of this article is structured as follows. In Section \ref{sec:methods}, we describe the proposed methods in detail. Section \ref{sec:experiments} outlines the experimental setup. Section \ref{sec:results} presents the empirical findings. Section \ref{sec:discussion} provides a critical analysis of the results, and Section \ref{sec:conclusion} concludes the work and discusses future directions.

\section{Related work}\label{sec:related}

\paragraph{Endoscopic Video Manifolds}
Early work on endoscopic video manifolds treated an entire procedure as a smooth trajectory in a low-dimensional space. Inter-frame similarities were defined using measures robust to blur, specular highlights, and turbid fluid, after which manifold learning methods such as Laplacian eigenmaps and Isomap-style embeddings were used to organise video frames \citep{Atasoy2012EndoscopicBiopsy}. These embeddings supported low-quality frame detection, scene clustering, and targeted biopsy guidance. However, they were non-generative: the learned manifolds could not inpaint pixels or reconstruct missing frames, did not provide probabilistic uncertainty, and were difficult to scale to modern endoscopic datasets.

\paragraph{VAEs and Generative Models}
Several studies have used Variational AutoEncoders (VAEs) or related generative models in endoscopy, usually for frame-level representation learning or image synthesis. An MMD-VAE was used for unsupervised tool-presence detection in surgical endoscopy videos, learning a low-dimensional representation that separates frames with and without tools without manual labels \citep{Li2020LearningSupervision}. A shared latent space between optical and virtual colonoscopy was also proposed using a lossy CycleGAN-like model, enabling visualisation of missed mucosal surfaces \citep{Mathew2021VisualizingRepresentations}. VAE-GAN hybrids have been used to synthesise endoscopic images, for example normal mucosa versus ulcers, to augment classification pipelines \citep{He2016DeepRecognition}. These approaches demonstrate the usefulness of compact latent spaces in endoscopy, but they largely treat frames as independent samples and emphasise representation, visualisation, or synthetic data generation.

More direct endoscopy-specific precedents are provided by VAE-based synthesis methods for wireless capsule endoscopy (WCE). EndoVAE proposed a VAE architecture for generating synthetic endoscopic images as an alternative to GAN-based augmentation, motivated by the instability, mode collapse, and data requirements of adversarial training in small medical datasets \citep{Diamantis2022EndoVAE:Autoencoder}. A later multiscale residual VAE framework incorporated multiscale convolutional feature extraction and residual connections to improve realism and diversity under limited-data conditions \citep{Diamantis2025MultiscaleSynthesis}. Both works focus on image synthesis for augmentation or dataset substitution. They do not address restoration of corrupted video segments or use predictive uncertainty as a reconstruction-quality signal.

A pre-trained Stable Diffusion VAE was used to encode endoscopy frames into a latent space, followed by a diffusion-based video generator and spatio-temporal transformer for realistic synthetic endoscopic videos and downstream 3D reconstruction \citep{Li2024Endora:Simulators}. Here, the latent space is powerful, but the generative prior is diffusion-based rather than Gaussian-process-based, and the VAE backbone is generic rather than endoscopy-specific. Related latent-space work in GI domains includes glottis segmentation in laryngeal endoscopy, where a single-channel bottleneck was sufficient for interpretable segmentation \citep{Dadras2024DeepFailures}, and a VAE trained on esophageal high-resolution manometry, where a four-dimensional latent space captured physiologically meaningful contraction patterns \citep{Kou2021AAutoencoder}. Collectively, these studies show that latent representations are useful in GI applications, but leave open the problem of uncertainty-aware restoration of degraded endoscopic video.

\paragraph{Endoscopic Foundation Models}
Large-scale endoscopic foundation models provide strong in-domain encoders and benchmarks. GastroNet-5M contains approximately five million images from around 500{,}000 procedures performed between 2012 and 2020 across eight hospitals in the Netherlands, spanning upper and lower GI endoscopy, multiple manufacturers, and diverse clinical indications \citep{Jong2026GastroNet-5M:Endoscopy}. Self-supervised pretraining of ResNet and Vision Transformer (ViT) backbones on this dataset using contrastive or DINO-style objectives improves performance over ImageNet-pretrained models across multiple GI classification and segmentation tasks, often with substantially fewer annotations \citep{Boers2024FoundationEfficiency}. This provides both pretrained encoders and a benchmark ecosystem that new endoscopic video methods should engage with.

EndoViT pretrains ViT architectures on Endo700k using a masked autoencoder objective, learning to reconstruct masked patches in endoscopy images and improving downstream segmentation and recognition compared with ImageNet-pretrained ViTs \citep{Batic2024EndoViT:Images}. Later work has extended this approach to federated settings \citep{Kirchner2026FederatedCollections}. These models remain deterministic representation learners, typically used as discriminative backbones rather than probabilistic latent-variable models.

Video-native foundation models have also emerged for endoscopic and surgical video analysis. Endo-FM pretrains a video transformer with masked reconstruction and contrastive objectives on approximately five million frames, achieving strong performance on downstream classification, segmentation, and detection tasks \citep{Wang2025ImprovingSequences,Wang2023FoundationPre-train, Jaspers2026ScalingModels}. MSN-style masked Siamese approaches \citep{Assran2022MaskedLearning}, UniSurg \citep{Shen2026UniSurg:Understanding}, and related JEPA-style methods learn strong clip-level representations by predicting masked latent features. In these models, temporal structure is captured implicitly through attention or self-supervised objectives, rather than through an explicit probabilistic prior over the latent trajectory.

\paragraph{Gaussian Process Prior VAE}
Gaussian Process Prior Variational Autoencoders (GPVAEs) \citep{Casale2018GaussianAutoencoders} extend standard VAEs by replacing the factorised Gaussian latent prior, with a Gaussian process prior indexed by time or other covariates. This allows correlations between latent variables to be modelled explicitly and supports principled interpolation, imputation, and uncertainty estimation. The original formulation, however, scales as $\mathcal{O}(N^3)$, which means cubically with the number of observations $N$. This is impractical for long endoscopic procedures.

Recent neighbor-driven GPVAE variants improve scalability by replacing the full covariance with local approximations \citep{Shi2025Neighbour-DrivenModelling}. In the Hierarchical Prior Approximation (HPA), each mini-batch is augmented with nearby points in the GP input space, restricting covariance computations to local subsets. In the Sparse Precision Approximation (SPA), the prior is factorised into low-order conditionals, so each latent variable depends only on a small neighbor set. After an initial nearest-neighbor search, training scales as $\mathcal{O}(D H^3 B)$ where \(D\) is the latent dimension, \(H\) the number of neighbors, and \(B\) the batch size. These methods perform competitively against full GPVAE \citep{Casale2018GaussianAutoencoders} and inducing-point SVGPVAE \citep{Jazbec2021Autoencoders} baselines on generic temporal and spatial datasets, but have not yet been applied to endoscopic video or combined with endoscopy-specific ViT encoders.

\section{Methods}\label{sec:methods}


A Gaussian process (GP) defines a distribution over functions rather than over finite-dimensional parameter vectors \citep{Rasmussen2006GaussianLearning}. A function \(f(\cdot)\) is said to follow a GP if any finite collection of function values has a joint Gaussian distribution:
\begin{equation}
f(\cdot) \sim \mathcal{GP}\big(m(\cdot), k(\cdot,\cdot)\big),
\end{equation}
where \(m(\cdot)\) is the mean function and \(k(\cdot,\cdot)\) is the covariance function or kernel. For inputs $\mathbf{X} = \{\mathbf{x}_1,\ldots,\mathbf{x}_N\}^\top$, the corresponding function values
\begin{equation}
\mathbf{f} = \big[f(\mathbf{x}_1), \ldots, f(\mathbf{x}_N)\big]^\top
\end{equation}
are distributed as
\begin{equation}
\mathbf{f} \sim \mathcal{N}(\boldsymbol{\mu}, \mathbf{K}),
\end{equation}
with
\begin{equation}
\boldsymbol{\mu}_i = m(\mathbf{x}_i),
\qquad
\mathbf{K}_{ij} = k(\mathbf{x}_i,\mathbf{x}_j).
\end{equation}
The kernel therefore determines which observations are expected to have similar function values. 

\begin{figure*}[!tbp] 
  \centering
  \includegraphics[width=\textwidth,height=0.82\textheight,keepaspectratio]{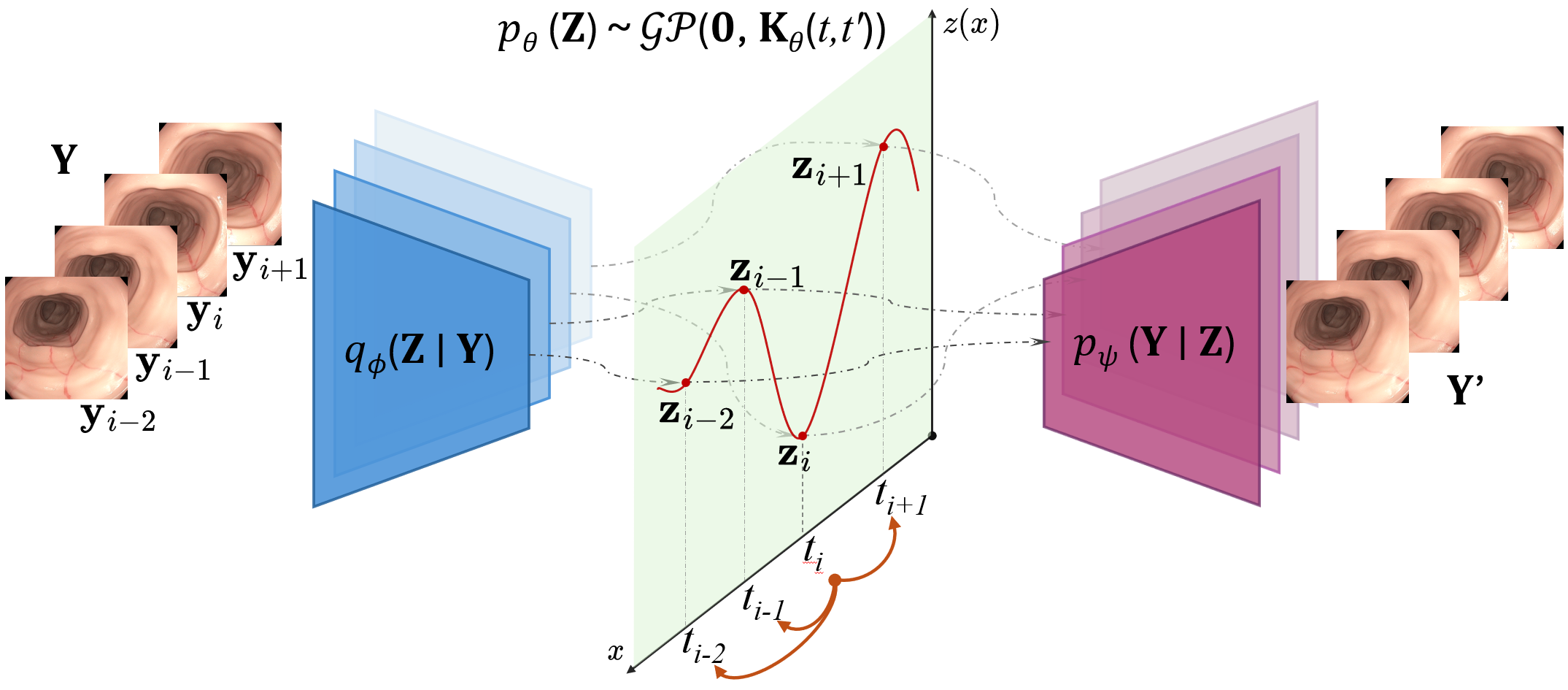}
  \caption{An overview of our proposed model. In a classical VAE setting, images $\mathbf{Y}$ are encoded by $q_{\phi}$ to latent points $\mathbf{Z}$. A decoder maps those to back to images $\mathbf{Y'}$. Additionally, we put a Gaussian process prior on the latent points, capturing covariances of neighboring frames in a video via their timestamps $t$.}
  \label{fig:overview}
\end{figure*}

In this work, the inputs $\mathbf{x}$ are the timestamps $t$ or equivalently frame numbers of the images. The resulting function values $\mathbf{f}$ of the Gaussian process are the points in the latent space $\mathbf{Z} = \{\mathbf{z}_1,\ldots,\mathbf{z}_N\}^\top$. A schematic overview of our model is given in Figure~\ref{fig:overview}.

This means that in our setting, instead of assuming independent latent variables $\mathbf{z}_{1:T}$ for each timestamp $t$,
\begin{equation}
p(\mathbf{z}) = \prod_{t=1}^{T} \mathcal{N}(z_t;0,I),
\end{equation}
we place a GP prior over each latent dimension $d$ across time:
\begin{equation}
z_{1:T}^{(d)} \sim \mathcal{GP}\big(0,k(t,t')\big),
\qquad d=1,\ldots,D.
\end{equation}
Equivalently, for a sequence of \(T\) frames,
\begin{equation}
\mathbf{z}^{(d)} =
\big[z_1^{(d)},\ldots,z_T^{(d)}\big]^\top
\sim
\mathcal{N}(\mathbf{0}, \mathbf{K}),
\end{equation}
where the entries of $\mathbf{K}$ are determined by the covariance function $k(t,t')$.

This covariance function or kernel specifies how strongly two function values are expected to co-vary. A common choice is the squared-exponential or radial basis function (RBF) kernel, which in general is given by:
\begin{equation}
k_{\mathrm{RBF}}(\mathbf{x},\mathbf{x}')
=
\sigma_f^2
\exp\left(
-\frac{\|\mathbf{x}-\mathbf{x}'\|^2}{2l^2}
\right),
\end{equation}
where \(\sigma_f^2\) is the signal variance and $l$ is the length-scale. The length-scale controls how quickly correlations decay with temporal distance. A large $l$ produces very smooth latent trajectories, while a small $l$ allows the latent variables to change more rapidly between frames. Sample paths of a Gaussian process with an RBF kernel are almost surely infinitely differentiable. This kernel is best suited when the underlying function is assumed to be very smooth \citep{Rasmussen2006GaussianLearning}.

Another useful choice is the Cauchy kernel:
\begin{equation}
k_{\mathrm{Cauchy}}(\mathbf{x},\mathbf{x}')
=
\sigma_f^2
\left(
1 + \frac{\|\mathbf{x}-\mathbf{x}'\|^2}{l^2}
\right)^{-\alpha},
\end{equation}
where $l$ again controls the temporal scale and \(\alpha > 0\) controls the heaviness of the kernel tail. Unlike the RBF kernel, it is much less smooth at the origin. While it is still differentiable, it allows for more wiggle or local variation compared to the rigid smoothness of RBF. The Cauchy kernel also decays more slowly with distance. This allows distant frames to remain weakly correlated, which can be useful in endoscopy when similar anatomical regions reappear after camera motion or when the video contains long-range temporal dependencies.

In a standard Gaussian process regression setting, training means learning the kernel hyperparameters, such as \(\sigma_f^2\), $l$, and \(\alpha\), from observed data. Suppose we observe noisy targets
\begin{equation}
\mathbf{y}
=
\mathbf{f}
+
\boldsymbol{\epsilon},
\qquad
\boldsymbol{\epsilon}
\sim
\mathcal{N}(\mathbf{0},\sigma_n^2\mathbf{I}),
\end{equation}
where \(\sigma_n^2\) is the observation-noise variance. The marginal distribution of the observations is
\begin{equation}
p(\mathbf{y}\mid \mathbf{x})
=
\mathcal{N}
\left(
\mathbf{y};
\mathbf{0},
\mathbf{K}_{\theta} + \sigma_n^2\mathbf{I}
\right),
\end{equation}
where \(\theta\) denotes the kernel hyperparameters. These hyperparameters are learned by maximizing the log marginal likelihood:
\begin{equation}
\begin{aligned}
\log p(\mathbf{y}\mid \mathbf{x},\theta)
={}&
-\frac{1}{2}
\mathbf{y}^{\top}
\left(
\mathbf{K}_{\theta} + \sigma_n^2 \mathbf{I}
\right)^{-1}
\mathbf{y}
\\
&-
\frac{1}{2}
\log
\left|
\mathbf{K}_{\theta} + \sigma_n^2 \mathbf{I}
\right|
\\
&-
\frac{T}{2}
\log(2\pi).
\end{aligned}
\end{equation}

The first term is a data fit term. The second term penalizes overly complex covariance structures. The third term is just a scalar value. In this way, Gaussian processes balance data fit and smoothness automatically. Consequently, Occam’s razor (the simplest explanation is usually the right one) is embedded within the model, inherently shielding the Gaussian process from overfitting \citep{Ras2000Occam}.

In a GPVAE, the Gaussian process is not trained directly on observed pixels. Instead, it acts as a prior over the latent variables inferred by a VAE encoder. Given an endoscopic video sequence \(\mathbf{y}_{1:T}\), the encoder defines an approximate posterior:
\begin{equation}
q_{\phi}(\mathbf{Z}\mid \mathbf{Y}),
\end{equation}
while the VAE decoder defines the likelihood:
\begin{equation}
p_{\psi}(\mathbf{Y'}\mid \mathbf{Z}),
\end{equation}
in which $\mathbf{Y'}$ are the reconstructed images.
The GP prior is given by
\begin{equation}
p_{\theta}(\mathbf{Z})
=
\prod_{d=1}^{D}
\mathcal{N}
\left(
\mathbf{z}^{(d)};
\mathbf{0},
\mathbf{K}_{\theta}
\right),
\end{equation}
where \(\mathbf{K}_{\theta}\) is computed from the chosen kernel.

In general, this model is trained by maximizing the evidence lower bound (ELBO):
\begin{equation}
\mathcal{L}
=
\mathbb{E}_{q_{\phi}(\mathbf{Z}\mid \mathbf{Y})}
\left[
\log p_{\psi}(\mathbf{Y}\mid \mathbf{Z})
\right]
-
\mathrm{KL}
\left[
q_{\phi}(\mathbf{Z}\mid \mathbf{Y})
\;\|\;
p_{\theta}(\mathbf{Z})
\right].
\end{equation}
The reconstruction term encourages the decoder to reproduce the observed frames, while the KL term encourages the latent variables to follow the temporal correlation structure specified by the GP prior. During training, the encoder parameters \(\phi\), decoder parameters \(\psi\), and kernel hyperparameters \(\theta\) can be optimised jointly by gradient descent. The practical training objective used in our implementation, including the reconstruction loss and the model-specific KL terms, is detailed in Appendix~\ref{app:objective}.

\section{Experiments}\label{sec:experiments}

\subsection{Models}\label{subsec:models}

In this work we compare the classic VAE to the GPVAE extensions for endoscopic videos. We implement the two approximations described in \citep{Shi2025Neighbour-DrivenModelling}, namely the Hierarchical Prior Approximation and the Sparse Precision Approximation (see also Section~\ref{sec:related}). Throughout, we used $H=50$ neighbors as a practical trade-off between temporal context and computational cost: increasing $H$ provides a richer local approximation to the GP prior, whereas smaller values reduce memory and runtime but may insufficiently capture smooth frame-to-frame dynamics. The exact KL terms used for the VAE, GPVAE-HPA, and GPVAE-SPA variants, together with the practical differences from \citep{Shi2025Neighbour-DrivenModelling}, are given in Appendix~\ref{app:kl_terms}.

We equip these models with several state-of-the-art encoders to assess how the latent prior interacts with different visual representations.

First, we use the convolutional encoder from EndoVAE \citep{Diamantis2022EndoVAE:Autoencoder}, which provides an endoscopy-specific autoencoding baseline. In this implementation, the encoder consists of a stack of strided convolutional layers with ReLU activations. Starting from the RGB input frame, the spatial resolution is halved at each layer using $4 \times 4$ convolutions with stride 2 and padding 1, while the number of feature channels is increased from 32 up to a maximum of 256. The resulting feature map is flattened and passed through a fully connected layer that outputs $2d_z$ values, which are split into the posterior mean and log-variance, $\mu_\phi(x)$ and $\log\sigma^2_\phi(x)$.

The EndoVAE decoder mirrors this structure. A latent sample $z$ is first projected with a fully connected layer to the bottleneck feature-map shape used by the encoder. The representation is then progressively upsampled using transposed convolutions with ReLU activations until the target image resolution is reached. A final $3 \times 3$ convolution maps the decoded feature map back to the RGB image space. For all VAE and GPVAE variants, we use this EndoVAE decoder unless stated otherwise, so that differences between model variants mainly reflect the encoder and latent-prior choices.

Second, we use an encoder based on a pretrained ViT from \citep{Boers2024FoundationEfficiency} as a feature extractor, allowing us to evaluate the effect of foundation-model features in the VAE and GPVAE settings. We also consider a low-rank adaptation strategy, where the same pretrained ViT is further finetuned, providing an alternative to the frozen or directly reused pretrained encoder. We refer to these encoders as GastroEnc and GaLoRAEnc respectively. For both variants, the input frame is resized to the ViT resolution of $256\times256$ and passed through the pretrained backbone, from which we use the resulting 768-dimensional image representation. This representation is mapped to the latent space by a lightweight multilayer perceptron head consisting of three hidden linear layers with GELU activations and dropout. Specifically, the head maps $768 \rightarrow 1024 \rightarrow 512 \rightarrow 512 \rightarrow 2d_z$, where $d_z$ is the latent dimensionality. The final output is split into the approximate posterior mean and log-variance, $\mu_\phi(x)$ and $\log\sigma^2_\phi(x)$.

Finally, we also compare to the multiscale vector-quantized VAE (MSVQ-VAE) proposed in \citep{Diamantis2025MultiscaleSynthesis}, which uses its own encoder--decoder architecture. In contrast to the continuous latent space used by the VAE and GPVAE models, MSVQ-VAE represents images through discrete codebook entries. 
In our implementation, the MSVQ-VAE encoder and decoder use three multiscale downsampling/upsampling stages. The vector quantizer uses a codebook with 256 entries, updated with exponential moving average updates using decay 0.99. The encoder output is quantized in a 64-dimensional code space, while an additional projection maps the spatial bottleneck representation to a 256-dimensional latent vector for compatibility with the rest of the latent-space analysis pipeline.

We train MSVQ-VAE with the same maximum number of epochs and base learning rate as the other models, namely 500 epochs and an Adam learning rate of $10^{-3}$. 
Because of the higher memory requirements of the multiscale encoder-decoder and vector-quantized bottleneck, we use a smaller batch size of 8 for MSVQ-VAE. 
The reconstruction objective uses an MSE weight of 5.0 and an LPIPS weight of 2.0, together with a VQ commitment weight of 0.25. These hyperparameters were selected empirically on endoscopic video reconstructions, prioritising a balance between pixel-level fidelity, perceptual sharpness, and stable vector-quantized codebook usage.

However, we do not combine MSVQ-VAE with the GPVAE prior, because the two models define fundamentally different latent spaces. 
The GPVAE assumes a continuous latent variable with a Gaussian approximate posterior, so that temporal correlations can be imposed through a Gaussian-process prior over the latent trajectories. MSVQ-VAE, in contrast, represents each image using discrete codebook indices at multiple spatial scales. 
These discrete quantized latents do not provide the continuous Gaussian posterior parameters required by the GPVAE objective, such as $\mu_\phi(x)$ and $\log\sigma^2_\phi(x)$. As a result, MSVQ-VAE is used as a separate reconstruction baseline rather than as an encoder-decoder backbone for the GPVAE variants.

An overview of the models implemented in this work is given in Table~\ref{tab:models}.


\begin{table*}[!tbp]
\caption{Overview of the evaluated VAE models and encoder-decoder combinations.}
\label{tab:models}
\begin{tabularx}{\textwidth}{@{}lllX@{}}
\toprule
Model & Encoder & Decoder & Summary\\
\midrule
VAE & EndoVAEEnc & EndoVAEDec & Baseline VAE with encoder and decoder as described in EndoVAE \citep{Diamantis2022EndoVAE:Autoencoder}\\
VAE & GastroEnc & EndoVAEDec & VAE using an encoder with pretrained ViT from \citep{Boers2024FoundationEfficiency} and decoder from EndoVAE \citep{Diamantis2022EndoVAE:Autoencoder}\\
VAE & GaLoRAEnc & EndoVAEDec & VAE using a low-rank adapted pretrained ViT from \citep{Boers2024FoundationEfficiency} and decoder from EndoVAE \citep{Diamantis2022EndoVAE:Autoencoder}\\
\midrule
GPVAE-HPA & EndoVAEEnc & EndoVAEDec & GPVAE with Hierarchical Prior Approximation \citep{Shi2025Neighbour-DrivenModelling}, encoder and decoder from EndoVAE \citep{Diamantis2022EndoVAE:Autoencoder}\\
GPVAE-HPA & GastroEnc & EndoVAEDec & GPVAE with Hierarchical Prior Approximation \citep{Shi2025Neighbour-DrivenModelling} using an encoder with pretrained ViT from \citep{Boers2024FoundationEfficiency}\\
GPVAE-HPA & GaLoRAEnc & EndoVAEDec & GPVAE with Hierarchical Prior Approximation \citep{Shi2025Neighbour-DrivenModelling} using a low-rank adapted pretrained ViT from \citep{Boers2024FoundationEfficiency}\\
\midrule
GPVAE-SPA & EndoVAEEnc & EndoVAEDec & GPVAE with Sparse Precision Approximation \citep{Shi2025Neighbour-DrivenModelling}, encoder and decoder from EndoVAE \citep{Diamantis2022EndoVAE:Autoencoder}\\
GPVAE-SPA & GastroEnc & EndoVAEDec & GPVAE with Sparse Precision Approximation \citep{Shi2025Neighbour-DrivenModelling} using an encoder with pretrained ViT from \citep{Boers2024FoundationEfficiency}\\
GPVAE-SPA & GaLoRAEnc & EndoVAEDec & GPVAE with Sparse Precision Approximation \citep{Shi2025Neighbour-DrivenModelling} using a low-rank adapted pretrained ViT from \citep{Boers2024FoundationEfficiency} \\
\midrule
MSVQ-VAE & MSVQEnc & MSVQDec & Multi-scale vector-quantized VAE baseline as described in \citep{Diamantis2025MultiscaleSynthesis}\\
\bottomrule
\end{tabularx}
\end{table*}

\subsection{Dataset}\label{subsec:dataset}

We evaluate our models on C3VDv2 \citep{Golhar2025C3VDv2ColonoscopyRealism}, a high-definition phantom-based colonoscopy video dataset designed for 3D reconstruction and endoscopic scene understanding. It includes two colon anatomies, divided into multiple anatomical segments and rendered with several texture/color variants, providing controlled but visually diverse endoscopic video sequences. C3VDv2 provides RGB colonoscopy frames together with rich geometric ground truth, including depth, surface normals, optical flow, occlusion maps, camera poses, coverage maps, and 3D models for the registered videos. The dataset also contains more challenging scenarios such as fecal debris, mucous pools, blood, lens obstruction, fast camera motion, en-face views, and deformation sequences. These properties make it suitable for evaluating temporally structured generative models on realistic endoscopic video data, while retaining access to frame order and camera-pose information.

To ensure computational feasibility during extensive baseline comparisons while still covering a broad range of visual and motion conditions, we selected a subset of 9 representative evaluation sequences from the 192 available videos. This subset was chosen to span anatomical locations, camera motions, sequence lengths, and visual artifacts. Specifically, the selected sequences include both clean and debris-containing environments across different colon segments, with challenging conditions such as polyps, blood, water on the lens, rapid oscillatory motion, and low-texture views. An overview of these curated evaluation sequences is given in Table~\ref{tab:c3vdv2_sequences}. A visualisation of frame 50 from each selected sequence is shown in Figure~\ref{fig:dataset_examples}.

\begin{table*}[!tbp]
\caption{Overview of the selected C3VDv2 video sequences with tags and comments as provided by \citep{Golhar2025C3VDv2ColonoscopyRealism}.}
\label{tab:c3vdv2_sequences}
\begin{tabularx}{\textwidth}{@{}l r r c X@{}}
\toprule
Video name & Frames & Camera speed & Debris & Tags / comments\\
\midrule
c1\_ascending\_t1\_v1 & 282 & 20 & no & Zigzag\\
c1\_descending\_t1\_v2 & 208 & 20 & no & Loop; saturation; back to start point\\
c1\_descending\_t2\_v3 & 617 & 10 & yes & Straight and oscillatory motion; blood\\
c1\_rectum\_t2\_v2 & 366 & 14 & no & Zigzag; exploratory motion\\
c1\_rectum\_t4\_v3 & 457 & 9 & yes & Fecal debris; withdrawal with oscillatory motion\\
c1\_transverse2\_t1\_v2 & 517 & 10 & no & Loop; polyp; inspect the polyp; start and end point is the same\\
c2\_cecum\_t3\_v2 & 306 & 30 & no & Fast; oscillatory motion in exploration\\
c2\_sigmoid\_t2\_v2 & 448 & 14 & no & Polyp; first and second half of the video have the same waypoints\\
c2\_transverse2\_t1\_v3 & 522 & 20 & yes &Cleaning polyp with water jet and scope dips into mucous pool which is cleaned with lens cleaning water.
\\
\bottomrule
\end{tabularx}
\renewcommand{\arraystretch}{1}
\end{table*}

All frames are resized to $336\times336$, converted to RGB and scaled from 8-bit intensity values to floating-point values $\tilde{x}_c \in [0,1]$. 
For the experiments reported here, we use the GastroNet channel normalization given in \citep{Boers2024FoundationEfficiency}. 
Each RGB channel is standardized independently as
\begin{equation}
x'_c = \frac{\tilde{x}_c - \mu_c}{\sigma_c},
\end{equation}
where $c \in \{R,G,B\}$ denotes the color channel, with
$\mu = (0.6404,\ 0.3613,\ 0.3133)$ and $\sigma = (0.1898,\ 0.1555,\ 0.1409)$. Using the same normalization for all encoder-decoder variants keeps the CNN-based models and the pretrained GastroNet models on a common input scale, while preserving compatibility with the preprocessing used for the pretrained encoder.

\subsection{Hyperparameters}\label{subsec:hypers}

All models were implemented in PyTorch and trained on a HPC cluster, using NVIDIA A100 GPU nodes. Gaussian-process components were implemented with GPyTorch. Unless stated otherwise, models were trained for a maximum of 500 epochs with mini-batches of 64 frames and early stopping with a patience of 25 epochs. Early stopping was only enabled after the first half of training, so that models were not stopped during the KL warm-up phase. The VAE and GPVAE variants used a latent dimensionality of $d_z=256$.

The hyperparameter values were chosen as stable default settings based on preliminary experiments on a subset of the data. In these initial runs, we prioritized robust convergence across all datasets and model families rather than tuning each configuration independently. The same settings were then kept fixed for the main experiments to ensure that differences between models were not driven by model-specific hyperparameter tuning.

For the VAE and GPVAE variants, the objective combines pixel-wise reconstruction, perceptual reconstruction, and a model-specific KL term. We used weights of 4.0 for MSE, 2.0 for LPIPS, and 1.0 for the KL term. MSE was computed at the full training resolution, while LPIPS was computed with a fixed VGG-based network after resizing images to $256 \times 256$. The full objective, including the masked reconstruction loss and the VAE-, HPA-, and SPA-specific KL terms, is given in Appendix~\ref{app:objective}.

Optimization was performed with Adam. The encoder head and decoder were trained with an initial learning rate of $10^{-3}$. For GaLoRA models, the low-rank adaptation parameters were trained with a learning rate of $10^{-5}$, rank $r=8$, and scaling factor $\alpha=8$. For GPVAE models, the GP kernel hyperparameters and likelihood noise were optimized jointly with the neural-network parameters using a learning rate ten times larger than the base learning rate. Gradients were clipped to a maximum norm of 1.0.

The learning rate was linearly warmed up from 10\% of its initial value to the full value during the first 50 epochs. It was then annealed with a cosine schedule for the remaining epochs, with a minimum learning rate of $10^{-5}$. The KL contribution was also annealed: $\beta_t$ was set to zero for the first 50 epochs and then linearly increased over the next 50 epochs.

For the GPVAE variants, we used the Cauchy kernel and $H=50$ nearest neighbors, computed in temporal input space using frame indices. For numerical stability, the GP hyperparameters were constrained during optimization. With unscaled frame indices, the lengthscale was constrained to $[1,30]$ frames, the outputscale to $[0.5,10]$, and the likelihood noise to $[10^{-3},1]$.

\subsection{Masking Bad Pixels}\label{sec:masking}

Specular reflections can negatively affect VAE training because they introduce bright, saturated image regions that are not well explained by the underlying anatomical scene. These highlights are often caused by the endoscope light source and viewing geometry rather than by stable tissue appearance. As a result, they can dominate pixel-wise reconstruction losses and encourage the model to allocate latent capacity to reproducing transient illumination artifacts instead of modelling the smoother anatomical and temporal structure of the video. This is particularly undesirable for GPVAE models, where the latent trajectory is expected to vary smoothly with frame position. To reduce this effect, we use a binary valid-pixel mask during training and evaluation in all experiments reported below.

\begin{figure*}[!tbp]
    \centering
    \includegraphics[width=\textwidth,height=0.82\textheight,keepaspectratio]{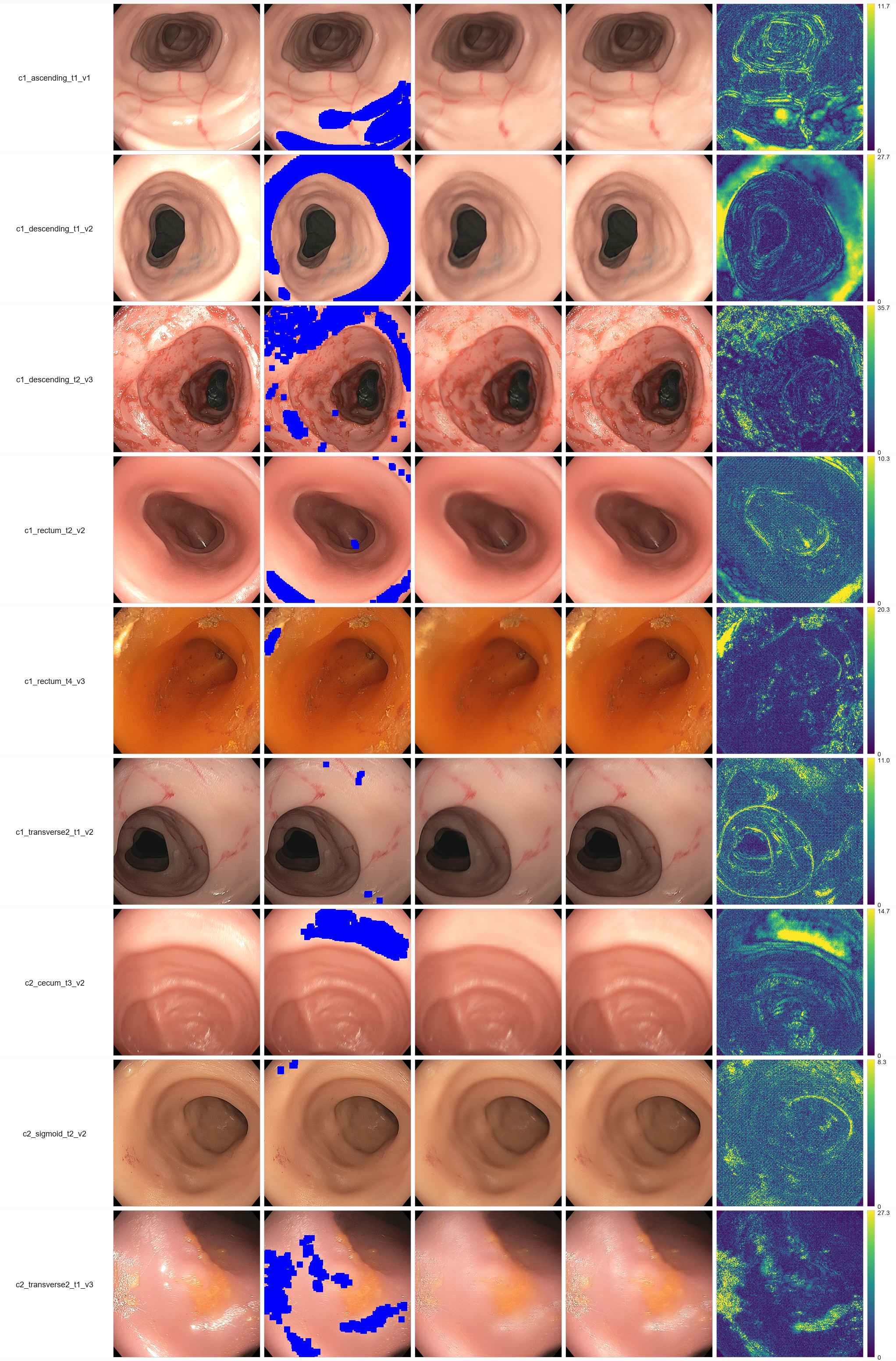}
    \caption{Qualitative reconstruction frame 50. From left to right: original frame, masked frame, VAE and GPVAE-HPA reconstruction, and difference between the VAE and GPVAE-HPA (independently scaled per sequence to the 98th percentile for visibility). Both reconstructions use the EndoVAEEnc encoder and EndoVAEDec decoder.} 
    \label{fig:dataset_examples}
\end{figure*}

To identify specular highlights, we use the DUCKNet-based specular segmentation model from the official implementation of \citep{Yu2026EndoscopicSegmentation}. Their method uses DUCKNet as the first stage of a two-stage endoscopic artifact inpainting pipeline, where DUCKNet localizes specular regions before mask-guided inpainting with LaMa and subsequent diffuse-artifact refinement with StableDelight. We use their released checkpoint as a fixed preprocessing model for generating specular-reflection masks. An example of a masked frame is given in Figure~\ref{fig:dataset_examples}, second column.

In our preprocessing pipeline, each frame is padded using reflected borders, resized to $256 \times 256$, normalized using ImageNet statistics, and passed through DUCKNet to obtain a single-channel specular probability map. Pixels with predicted probability above 0.1 are marked as artefacts. Because saturated highlights may still be missed by the learned model, we additionally mark pixels with grayscale intensity above 225 as invalid. This threshold was selected empirically as a conservative high-intensity cutoff on the 8-bit grayscale scale, intended to capture near-saturated highlights missed by DUCKNet while avoiding excessive masking of normal bright mucosal regions. The DUCKNet prediction and intensity-threshold mask are combined, dilated using a $5 \times 5$ kernel for two iterations, resized back to the original image resolution with nearest-neighbor interpolation, and cropped to remove the padding. The resulting artefact mask is inverted to obtain a valid-pixel mask, where valid anatomical pixels have value 1 and detected specular or overexposed regions have value 0. This mask is stored with the prepared dataset and used during training and evaluation to prevent transient reflection artefacts from dominating the reconstruction objective. In the experiments, the same valid-pixel mask defines the pixel set $\Omega$ used in the masked reconstruction loss described in Appendix~\ref{app:reconstruction_objective}.

\section{Results}\label{sec:results}

\subsection{Reconstruction}\label{sec:recon}

In this first experiment we provide the models with all frames in the videos and measure how well the encoder-decoder setup can reconstruct the given images. Pixel-level reconstruction accuracy is measured using masked root mean squared error (RMSE) over the valid-pixel mask. We also report peak signal-to-noise ratio (PSNR), computed from the same denormalized images with data range 1.0. Perceptual and structural image quality are assessed using SSIM and FID. SSIM is computed between reconstructed and real frames in denormalized RGB space. We compute an SSIM map per frame and average only over valid mask pixels. FID is computed with the TorchMetrics Frechet Inception Distance implementation using 2048-dimensional Inception features. Before FID computation, images are clipped to $[0,1]$, converted to 8-bit RGB, and compared as real versus reconstructed image sets. 

All metrics are computed after denormalization, i.e. after multiplying each channel by its standard deviation and adding the corresponding mean. This ensures that the reported values are measured in the original RGB image space rather than in the standardized network-input space. RMSE, SSIM, and PSNR are evaluated over the valid pixels defined by the mask, whereas FID is computed on the full clipped RGB frames. SSIM should be interpreted with caution because the reconstructions are already extremely close in pixel space, causing the score to saturate near 1 and making meaningful differences between models difficult to resolve. FID should also be treated cautiously because it is computed on a relatively small and domain-specific set of video frames, where Inception features may not reliably reflect perceptual quality in endoscopic imagery. We include them for completeness.

The results are given in Table~\ref{tab:image_encoded_metrics}. For the GPVAE, we also included the metrics for images recreated purely from their timestamp, instead of from the original image. Those results are given in Table~\ref{tab:time_encoded_metrics}. The non-GPVAE models, i.e. VAE and MSVQ-VAE, lack this ability.

\begin{table*}[!tbp]
\caption{Image-encoded reconstruction performance averaged across videos. Values are reported as mean $\pm$ population standard deviation. RMSE is measured in denormalized image intensity units on the $[0,1]$ scale, PSNR in dB, SSIM and FID are unitless, and time is reported as seconds per epoch. The best value in each metric column within each section is shown in bold.}
\label{tab:image_encoded_metrics}
\centering
\scriptsize
\setlength{\tabcolsep}{2pt}
\renewcommand{\arraystretch}{1.12}
\resizebox{\textwidth}{!}{%
\begin{tabular}{@{}lllccccc@{}}
\toprule
Model & Encoder & Decoder & RMSE $[0,1]$ $\downarrow$ & PSNR [dB] $\uparrow$ & SSIM [unitless] $\uparrow$ & FID [unitless] $\downarrow$ & Time/epoch [s] $\downarrow$ \\
\midrule
VAE & EndoVAEEnc & EndoVAEDec & $0.0023 \pm 0.0009$ & $53.41 \pm 3.25$ & $0.99994 \pm 0.00005$ & $0.153 \pm 0.114$ & $\mathbf{2.62 \pm 0.80}$ \\
GPVAE\_HPA & EndoVAEEnc & EndoVAEDec & $\mathbf{0.0017 \pm 0.0006}$ & $\mathbf{55.79 \pm 2.85}$ & $\mathbf{0.99997 \pm 0.00005}$ & $\mathbf{0.080 \pm 0.100}$ & $3.06 \pm 0.88$ \\
GPVAE\_SPA & EndoVAEEnc & EndoVAEDec & $0.0018 \pm 0.0007$ & $55.25 \pm 3.12$ & $\mathbf{0.99997 \pm 0.00005}$ & $0.090 \pm 0.098$ & $2.93 \pm 0.84$ \\
\midrule
VAE & GastroEnc & EndoVAEDec & $0.0025 \pm 0.0010$ & $52.82 \pm 3.21$ & $0.99992 \pm 0.00008$ & $0.167 \pm 0.125$ & $\mathbf{2.48 \pm 0.75}$ \\
GPVAE\_HPA & GastroEnc & EndoVAEDec & $0.0022 \pm 0.0009$ & $53.57 \pm 3.17$ & $0.99989 \pm 0.00009$ & $0.136 \pm 0.110$ & $2.72 \pm 0.80$ \\
GPVAE\_SPA & GastroEnc & EndoVAEDec & $\mathbf{0.0021 \pm 0.0008}$ & $\mathbf{53.96 \pm 3.17}$ & $\mathbf{0.99994 \pm 0.00005}$ & $\mathbf{0.125 \pm 0.113}$ & $2.61 \pm 0.76$ \\
\midrule
VAE & GaLoRAEnc & EndoVAEDec & $0.0025 \pm 0.0010$ & $52.70 \pm 3.19$ & $0.99992 \pm 0.00008$ & $0.169 \pm 0.129$ & $\mathbf{9.46 \pm 2.88}$ \\
GPVAE\_HPA & GaLoRAEnc & EndoVAEDec & $0.0021 \pm 0.0007$ & $54.15 \pm 2.75$ & $0.99993 \pm 0.00005$ & $0.112 \pm 0.130$ & $12.99 \pm 3.89$ \\
GPVAE\_SPA & GaLoRAEnc & EndoVAEDec & $\mathbf{0.0019 \pm 0.0007}$ & $\mathbf{54.77 \pm 3.12}$ & $\mathbf{0.99996 \pm 0.00005}$ & $\mathbf{0.111 \pm 0.107}$ & $12.86 \pm 3.85$ \\
\midrule
MSVQ\_VAE & MSVQEnc & MSVQDec & $0.0023 \pm 0.0011$ & $53.44 \pm 3.69$ & $0.99992 \pm 0.00008$ & $0.133 \pm 0.113$ & $7.41 \pm 2.24$ \\
\bottomrule
\end{tabular}%
}%
\renewcommand{\arraystretch}{1}
\end{table*}

\begin{table*}[!tbp]
\caption{Time-encoded reconstruction performance averaged across videos. Values are reported as mean $\pm$ population standard deviation. RMSE is measured in denormalized image intensity units on the $[0,1]$ scale, PSNR in dB, and SSIM and FID are unitless. The best value in each metric column is shown in bold.}
\label{tab:time_encoded_metrics}
\centering
\scriptsize
\setlength{\tabcolsep}{4pt}
\resizebox{\textwidth}{!}{%
\begin{tabular}{@{}lllcccc@{}}
\toprule
Model & Encoder & Decoder & RMSE $[0,1]$ $\downarrow$ & PSNR [dB] $\uparrow$ & SSIM [unitless] $\uparrow$ & FID [unitless] $\downarrow$ \\
\midrule
GPVAE\_HPA & EndoVAEEnc & EndoVAEDec & $\mathbf{0.0017 \pm 0.0006}$ & $\mathbf{55.71 \pm 2.88}$ & $\mathbf{0.99997 \pm 0.00005}$ & $\mathbf{0.081 \pm 0.099}$ \\
GPVAE\_SPA & EndoVAEEnc & EndoVAEDec & $0.0018 \pm 0.0007$ & $55.23 \pm 3.12$ & $\mathbf{0.99997 \pm 0.00005}$ & $0.091 \pm 0.098$ \\
GPVAE\_HPA & GastroEnc & EndoVAEDec & $0.0023 \pm 0.0009$ & $53.53 \pm 3.18$ & $0.99989 \pm 0.00009$ & $0.137 \pm 0.109$ \\
GPVAE\_SPA & GastroEnc & EndoVAEDec & $0.0021 \pm 0.0009$ & $54.01 \pm 3.21$ & $0.99994 \pm 0.00005$ & $0.124 \pm 0.114$ \\
GPVAE\_HPA & GaLoRAEnc & EndoVAEDec & $0.0021 \pm 0.0007$ & $54.08 \pm 2.77$ & $0.99993 \pm 0.00005$ & $0.112 \pm 0.130$ \\
GPVAE\_SPA & GaLoRAEnc & EndoVAEDec & $0.0019 \pm 0.0007$ & $54.79 \pm 3.16$ & $0.99996 \pm 0.00005$ & $0.112 \pm 0.108$ \\
\bottomrule
\end{tabular}%
}%
\renewcommand{\arraystretch}{1}
\end{table*}

From the reconstruction tables, the clearest result is that adding the GP prior improves reconstruction quality most strongly when using the EndoVAE encoder-decoder. In the image-encoded setting, GPVAE-HPA with EndoVAEEnc and EndoVAEDec obtains the best RMSE, PSNR, SSIM, and FID, indicating that the temporal prior improves reconstruction even when the image itself is available to the encoder. Across the three matched encoder--decoder groups, selecting the better GPVAE approximation within each group reduced the mean image-encoded RMSE from 0.0024 for the VAE baselines to 0.0019, corresponding to a relative reduction of 21.9\%. The largest reduction was observed for EndoVAEEnc--EndoVAEDec, where GPVAE-HPA reduced RMSE from 0.0023 to 0.0017, a 26.1\% improvement.

The time-encoded table shows that the GPVAE models can reconstruct frames from timestamp alone with only a small loss compared with image-encoded reconstruction. This is important: it suggests that the learned latent trajectory captures meaningful temporal structure in the video, rather than merely acting as a regularizer during image reconstruction.

Across GPVAE variants, HPA performs best with the EndoVAE encoder, while SPA is slightly better for the GastroEnc and GaLoRAEnc variants in several metrics. This suggests that the best GP approximation may depend on the encoder representation, with no single variant dominating across all backbones.

The pretrained GastroNet-based encoders do not clearly outperform the simpler EndoVAE encoder in this reconstruction setting. In fact, EndoVAEEnc combined with GPVAE-HPA gives the strongest overall results, while GastroEnc and GaLoRAEnc generally produce higher RMSE and lower PSNR. This may indicate that the pretrained encoders provide useful semantic features, but are less optimized for precise pixel-level reconstruction than the dedicated convolutional encoder.

Finally, GaLoRAEnc substantially increases training time per epoch, especially for GPVAE models, without producing a corresponding improvement in reconstruction quality. For this experiment, the extra adaptation cost does not appear justified by the reconstruction metrics.

To illustrate this further, we provide a qualitative comparison in Figure~\ref{fig:dataset_examples}. The VAEs without a Gaussian process prior tend to yield more blurry images.

\subsection{Missing Frames}\label{subsec:frames}

To investigate the frame reconstruction performance, we compared experiments trained with test fractions of 10\%, 20\%, 40\% and 60\%. For every model and encoder/decoder combination, the reported test RMSE values were extracted and averaged across the videos.

Three quantities were considered. First, the RMSE was evaluated for all frames in the test set for all available models. Second, the time-encoded test RMSE was evaluated only for the GPVAE models, since this metric is only defined for models with a time-encoded reconstruction capability. Third, for the GPVAE models, we evaluated the Gaussian process posterior uncertainty of the time-encoded frames. This uncertainty was obtained from the per-frame uncertainty, excluding the time points used to train the Gaussian process.

\begin{figure*}[!tbp]
    \centering
    \includegraphics[width=\textwidth,height=0.82\textheight,keepaspectratio]{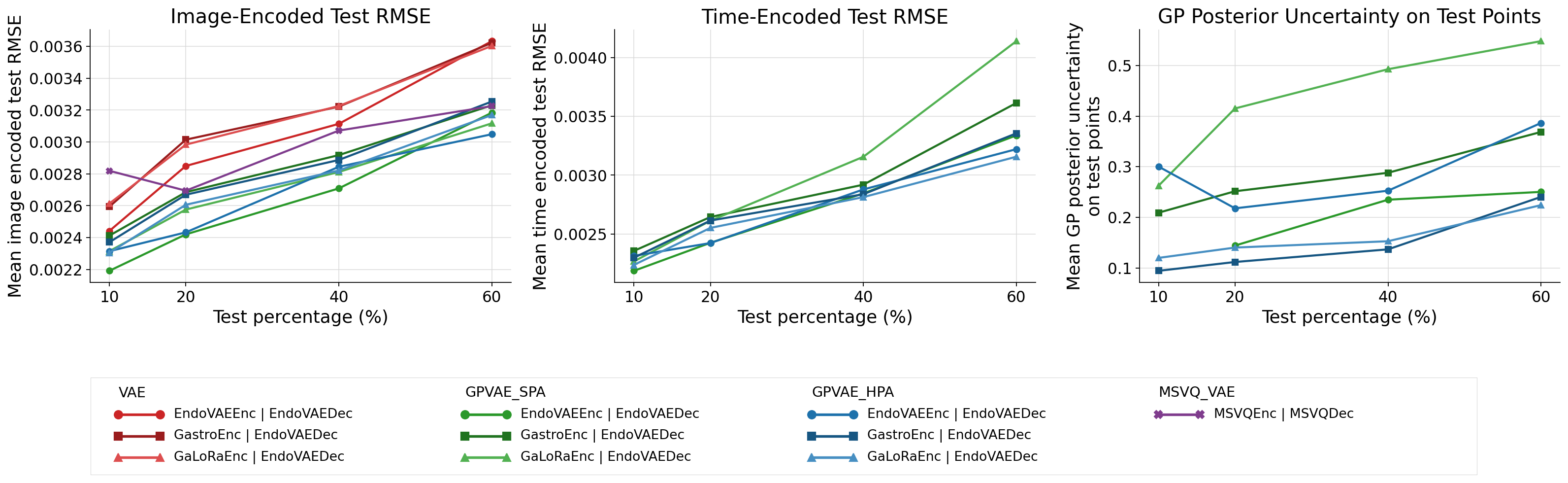}
    \caption{
    Effect of the test split percentage on reconstruction performance and Gaussian-process uncertainty.
    The left panel shows the image-encoded test RMSE averaged across datasets for all models and encoder/decoder combinations.
    The middle panel shows the time-encoded test RMSE for the GPVAE models.
    The right panel shows the mean Gaussian-process posterior uncertainty on test points for the GPVAE models.
    }
    \label{fig:test_percentage_rmse_uncertainty}
\end{figure*}

The GPVAE models show broadly similar behavior in the image-encoded and time-encoded RMSE curves. Configurations using the EndoVAE encoder generally remain among the strongest performers, while the relative ranking between encoders depends on the model variant and test fraction. At the 60\% test split, most configurations show a clear increase in RMSE, consistent with the expected degradation when fewer frames are available for training.

The Gaussian-process posterior uncertainty provides a complementary view of this effect. Since the uncertainty was averaged only over test points, it reflects the model's confidence at held-out time points rather than at the time points used for GP training. The uncertainty curves show that the GPVAE configurations do not respond uniformly to the test split percentage: some configurations show increased uncertainty at larger test fractions, while others remain relatively stable or decrease. This suggests that reconstruction error and posterior uncertainty capture related but distinct aspects of model behavior.

Unlike the RMSE, the GP posterior uncertainty does not increase monotonically with the test split percentage. This is expected, since the posterior uncertainty depends primarily on the temporal location of the held-out points relative to the GP training points and on the learned kernel hyperparameters, rather than directly on the image reconstruction error. Consequently, the GP can remain confident at test points that are well supported by nearby training points, even when reconstruction accuracy decreases due to reduced training data. The Gaussian process posterior uncertainty provides an additional diagnostic beyond reconstruction error alone.

\subsection{Camera Path Reconstruction}\label{subsec:path}

Image reconstruction quality does not necessarily translate to downstream tasks such as navigation or $3$D reconstruction. In endoscopy, this distinction is particularly important because low texture, specular highlights, blur, and repetitive tubular anatomy can make camera-motion estimation unstable even when individual frames look visually plausible. We therefore complement the image reconstruction evaluation with a camera path reconstruction analysis.

For each sequence in Table~\ref{tab:c3vdv2_sequences} we investigated, the ground-truth camera trajectory is provided. To assess how well reconstructed sequences preserve motion information, we estimate camera paths using both a classical visual odometry (VO) pipeline and a deep-learning-based pose estimator PoseNet \citep{Kendall2015PoseNet:Relocalization}, and compare the resulting trajectories against the ground-truth path.

The deep-learning-based PoseNet approach replaces explicit geometric estimation with a learned mapping from image pairs to relative camera motion. The model follows the standard self-supervised monocular pose-estimation design used in depth-and-pose learning. Two RGB frames are concatenated channel-wise and passed through a convolutional pose encoder, which produces a learned feature representation of the pair. A pose decoder then regresses a $6$-DoF relative motion parametrization consisting of axis-angle rotation and translation. In our experiments, we use the PoseNet implementation provided by SHADeS \citep{Daher2025SHADeS:Decomposition}, which was pretrained on endoscopic images.

The global trajectory is recovered by recursively composing the estimated relative motions. In the monocular case, the resulting trajectory is only determined up to an unknown global scale. Moreover, its quality depends strongly on the availability of stable and distinctive inter-frame motion cues. This makes trajectory estimation particularly sensitive in endoscopy, where specular highlights, low texture, repeated folds, blur, and forward motion can weaken geometric constraints.

The two trajectory estimators, VO and PoseNet, probe different aspects of reconstruction quality. Classical VO evaluates whether the reconstructed frames preserve sufficient geometric detail and correspondence structure for feature-based motion recovery. The pretrained SHADeS PoseNet instead evaluates whether the reconstructed sequence preserves the higher-level temporal cues required by a learned relative-pose model. Agreement between the two suggests that the restored video remains geometrically and temporally coherent, whereas divergence between them can help diagnose whether failure arises from poor feature geometry, poor learned motion cues, or both.

Because monocular trajectory estimation is only defined up to an arbitrary global scale, rotation, and translation, the estimated paths cannot be compared directly to the ground-truth trajectory in their raw form. We align each estimated trajectory to the ground truth using an in-house implementation of a Procrustes-style similarity alignment based on singular value decomposition \citep{Umeyama1991Least-SquaresPatterns}. This alignment removes trivial coordinate-system differences and allows the evaluation to focus on the actual geometric agreement between the trajectory shapes rather than on arbitrary global offsets.

After aligning each estimated trajectory to the ground-truth path, we evaluate camera path reconstruction using three complementary metrics: root mean squared error (RMSE), absolute trajectory error (ATE), and relative pose error (RPE). More details are given in Appendix~\ref{app:pathrecon}. Together, these metrics allow us to distinguish between global trajectory mismatch and local motion inconsistency.

\begin{figure*}[!tbp]
    \centering
    \includegraphics[width=\textwidth,height=0.82\textheight,keepaspectratio]{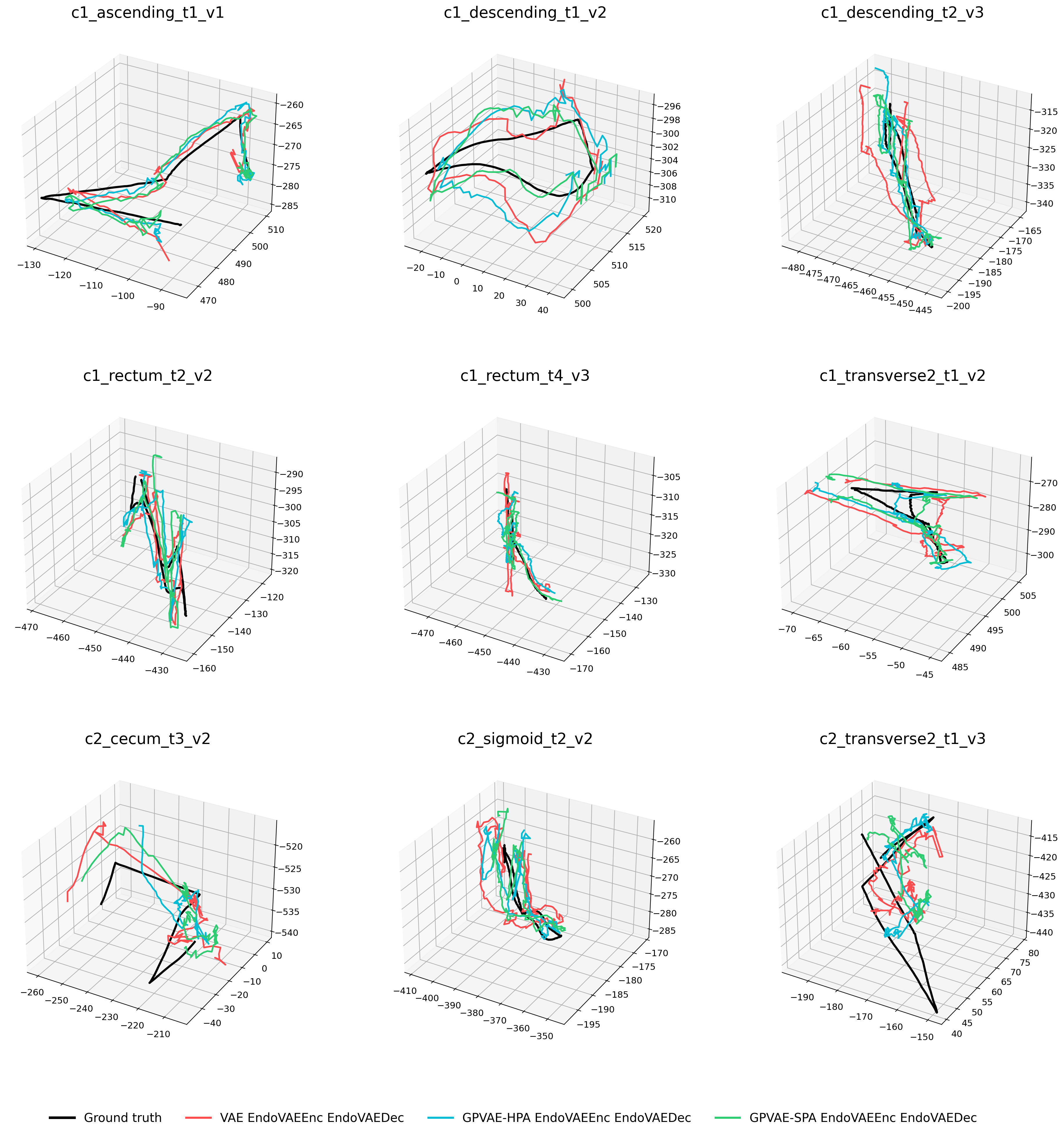}
    \caption{Aligned trajectory reconstructions obtained with classical visual odometry for the selected VAE, GPVAE-HPA, and GPVAE-SPA models across all evaluated videos. In each subplot, the reconstructed trajectories are shown together with the corresponding ground-truth camera path.}
    \label{fig:all_datasets_3x3_classical_vo_vs_ground_truth}
\end{figure*}

\begin{figure*}[!tbp]
    \centering
    \includegraphics[width=\textwidth,height=0.82\textheight,keepaspectratio]{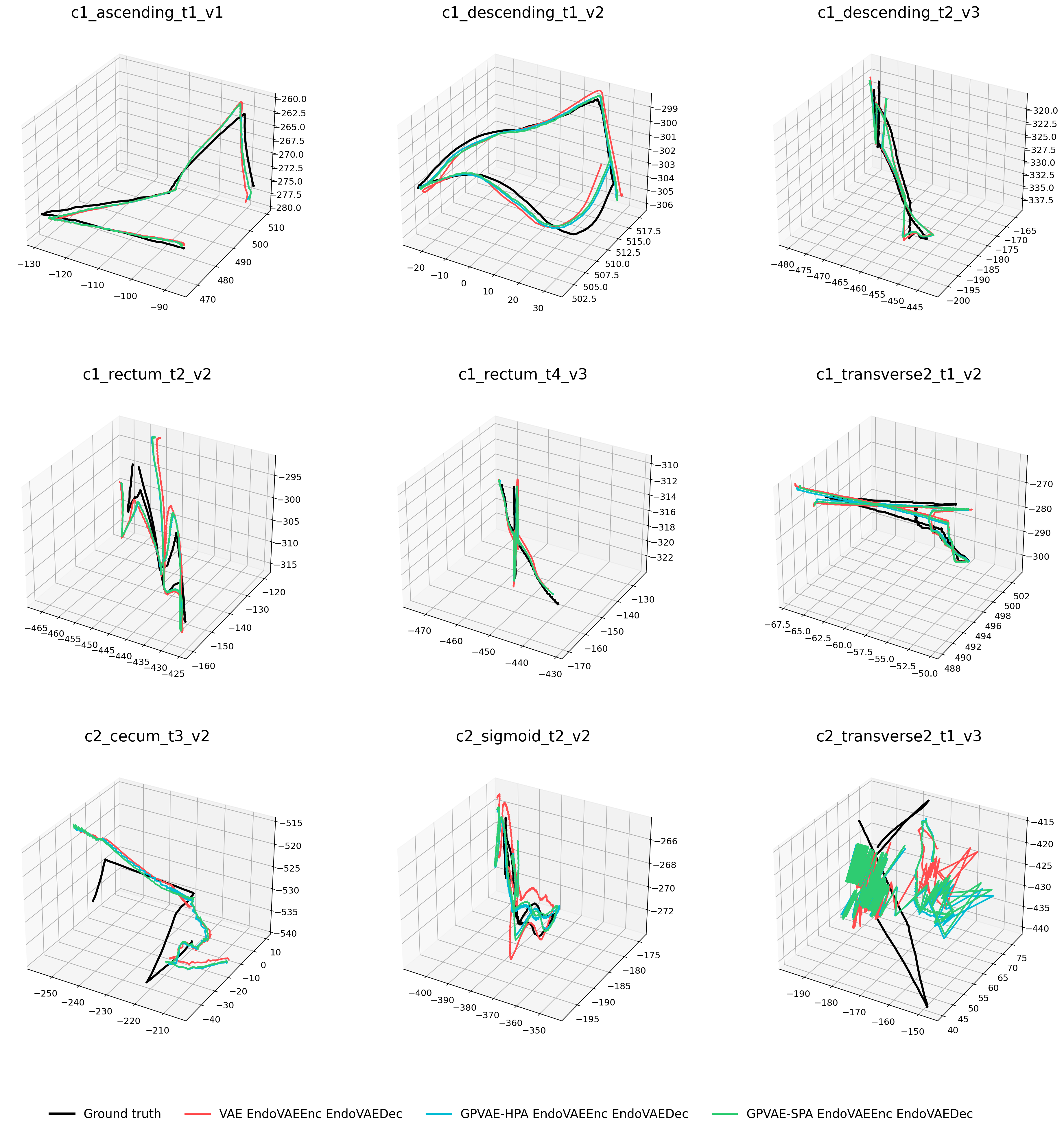}
    \caption{Aligned trajectory reconstructions obtained with the pretrained PoseNet for the selected VAE, GPVAE-HPA, and GPVAE-SPA models across all evaluated videos. In each subplot, the reconstructed trajectories are shown together with the corresponding ground-truth camera path.}
    \label{fig:all_datasets_3x3_learned_pose_vs_ground_truth}
\end{figure*}

\begin{figure*}[!tbp]
    \centering
    \includegraphics[width=\textwidth,height=0.82\textheight,keepaspectratio]{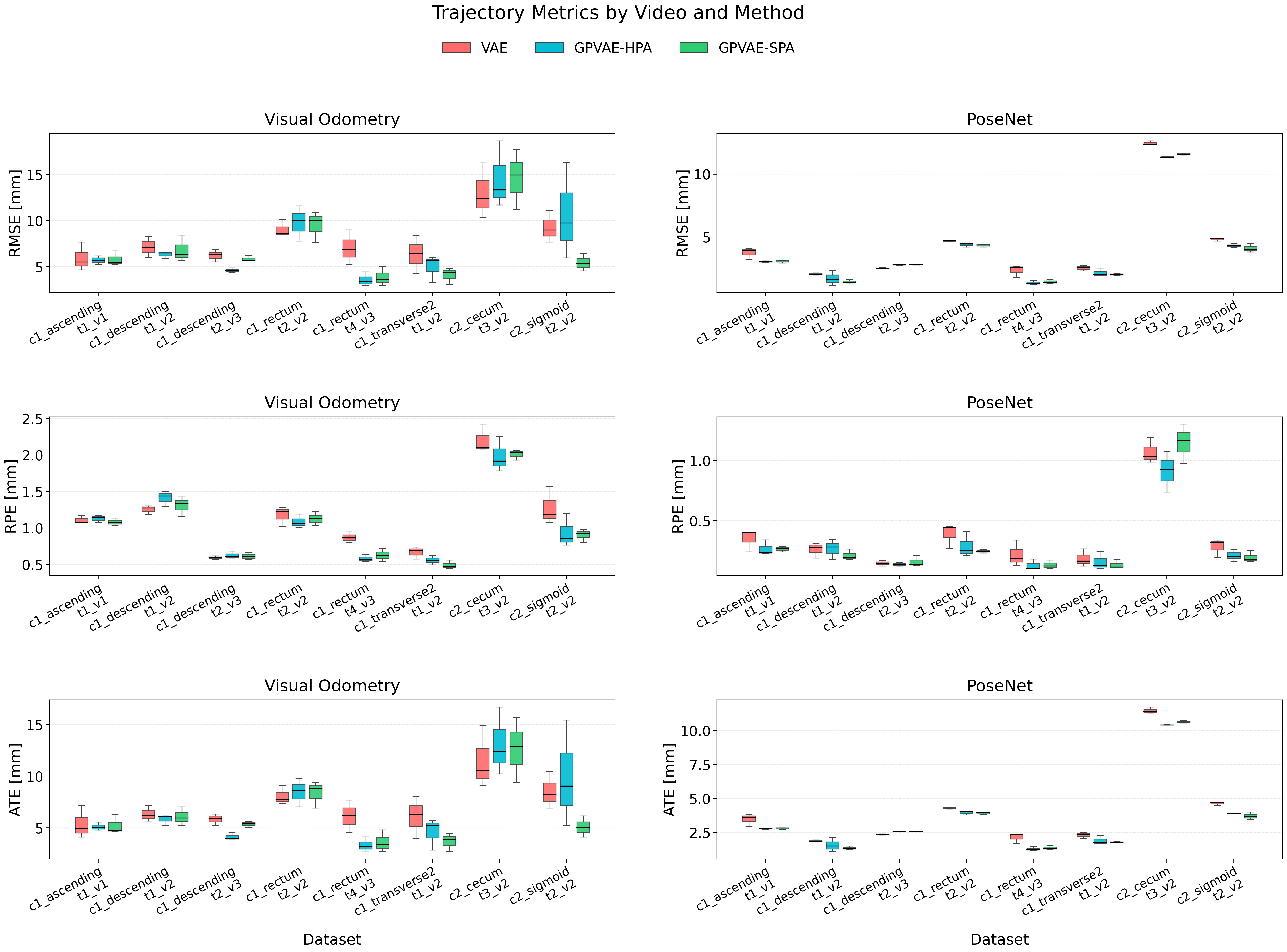}
    \caption{Distribution of trajectory reconstruction errors across videos for the VAE, GPVAE-HPA, and GPVAE-SPA model families, evaluated with visual odometry and PoseNet. Lower values are better. Within each family, the spread reflects the different encoder-decoder model combinations.}
    \label{fig:trajectory_metrics_boxplots_by_video_and_method}
\end{figure*}

Figures~\ref{fig:all_datasets_3x3_classical_vo_vs_ground_truth} and \ref{fig:all_datasets_3x3_learned_pose_vs_ground_truth} provide qualitative trajectory comparisons for the classical visual odometry and PoseNet-based reconstructions, respectively. In each case, the reconstructed trajectories of the selected VAE and GPVAE variants are aligned to the ground-truth path, allowing visual assessment of global drift, local deviations, and overall path fidelity across the different videos. Figure~\ref{fig:trajectory_metrics_boxplots_by_video_and_method} complements these qualitative examples with a quantitative summary, showing the distribution of RMSE, RPE, and ATE across videos for the VAE, GPVAE-HPA, and GPVAE-SPA model families under both trajectory estimation methods. We excluded video \texttt{c2\_transverse\_t1\_v3} from this analysis, as its trajectory estimates were not meaningful and no valid camera path could be reconstructed, which would otherwise dominate and clutter the visualisation. Table~\ref{tab:path_reconstruction_metrics} summarizes the path reconstruction performance averaged across videos for both classical visual odometry and PoseNet, and shows that the GPVAE variants generally improve trajectory preservation relative to the plain VAE baseline. Averaged across matched encoder--decoder groups and across both trajectory estimators, selecting the better GPVAE approximation reduced trajectory RMSE by 12.7\% relative to the corresponding VAE baselines.

\begin{table*}[!tbp]
\caption{Path reconstruction performance averaged across videos, excluding \texttt{c2\_transverse2\_t1\_v3} because its trajectory reconstruction results are severely degraded and would disproportionately skew the summary. Values are reported as mean $\pm$ population standard deviation. RMSE, ATE, and RPE are measured in mm. Lower values are better. The best value in each metric column within each encoder block and method section is shown in bold.}
\label{tab:path_reconstruction_metrics}
\centering
\scriptsize
\setlength{\tabcolsep}{5pt}
\renewcommand{\arraystretch}{1.08}
\begin{tabularx}{\textwidth}{
  >{\raggedright\arraybackslash}p{0.13\textwidth}
  >{\raggedright\arraybackslash}p{0.12\textwidth}
  >{\raggedright\arraybackslash}p{0.12\textwidth}
  >{\centering\arraybackslash}X
  >{\centering\arraybackslash}X
  >{\centering\arraybackslash}X}
\toprule
Model & Encoder & Decoder & RMSE [mm] $\downarrow$ & ATE [mm] $\downarrow$ & RPE [mm] $\downarrow$ \\
\midrule
\multicolumn{6}{l}{\textbf{Visual Odometry}} \\
VAE & EndoVAEEnc & EndoVAEDec & $6.63 \pm 1.90$ & $5.96 \pm 1.60$ & $1.08 \pm 0.46$ \\
GPVAE\_HPA & EndoVAEEnc & EndoVAEDec & $6.86 \pm 4.70$ & $6.17 \pm 4.20$ & $\mathbf{1.01 \pm 0.44}$ \\
GPVAE\_SPA & EndoVAEEnc & EndoVAEDec & $\mathbf{6.39 \pm 3.50}$ & $\mathbf{5.69 \pm 2.98}$ & $1.04 \pm 0.47$ \\
\midrule
VAE & GastroEnc & EndoVAEDec & $8.10 \pm 1.83$ & $7.33 \pm 1.41$ & $1.18 \pm 0.53$ \\
GPVAE\_HPA & GastroEnc & EndoVAEDec & $7.64 \pm 2.75$ & $6.89 \pm 2.26$ & $1.11 \pm 0.52$ \\
GPVAE\_SPA & GastroEnc & EndoVAEDec & $\mathbf{6.97 \pm 2.24}$ & $\mathbf{6.29 \pm 1.78}$ & $\mathbf{1.05 \pm 0.48}$ \\
\midrule
VAE & GaLoRAEnc & EndoVAEDec & $9.27 \pm 3.25$ & $8.41 \pm 3.03$ & $1.18 \pm 0.44$ \\
GPVAE\_HPA & GaLoRAEnc & EndoVAEDec & $8.06 \pm 4.35$ & $7.36 \pm 4.11$ & $1.00 \pm 0.43$ \\
GPVAE\_SPA & GaLoRAEnc & EndoVAEDec & $\mathbf{7.63 \pm 4.44}$ & $\mathbf{6.79 \pm 3.83}$ & $\mathbf{0.97 \pm 0.44}$ \\
\midrule
MSVQ\_VAE & MSVQEnc & MSVQDec & $5.78 \pm 1.58$ & $5.17 \pm 1.33$ & $1.16 \pm 0.48$ \\
\midrule
\multicolumn{6}{l}{\textbf{PoseNet}} \\
VAE & EndoVAEEnc & EndoVAEDec & $4.24 \pm 3.34$ & $3.92 \pm 3.12$ & $0.31 \pm 0.34$ \\
GPVAE\_HPA & EndoVAEEnc & EndoVAEDec & $\mathbf{3.78 \pm 3.10}$ & $\mathbf{3.42 \pm 2.84}$ & $\mathbf{0.29 \pm 0.30}$ \\
GPVAE\_SPA & EndoVAEEnc & EndoVAEDec & $3.83 \pm 3.13$ & $3.49 \pm 2.87$ & $0.32 \pm 0.38$ \\
\midrule
VAE & GastroEnc & EndoVAEDec & $4.46 \pm 3.15$ & $4.12 \pm 2.92$ & $0.41 \pm 0.23$ \\
GPVAE\_HPA & GastroEnc & EndoVAEDec & $3.94 \pm 2.95$ & $3.62 \pm 2.72$ & $\mathbf{0.33 \pm 0.17}$ \\
GPVAE\_SPA & GastroEnc & EndoVAEDec & $\mathbf{3.92 \pm 3.08}$ & $\mathbf{3.58 \pm 2.84}$ & $0.35 \pm 0.31$ \\
\midrule
VAE & GaLoRAEnc & EndoVAEDec & $4.51 \pm 3.12$ & $4.16 \pm 2.87$ & $0.37 \pm 0.27$ \\
GPVAE\_HPA & GaLoRAEnc & EndoVAEDec & $3.95 \pm 2.96$ & $3.60 \pm 2.73$ & $0.27 \pm 0.26$ \\
GPVAE\_SPA & GaLoRAEnc & EndoVAEDec & $\mathbf{3.81 \pm 3.08}$ & $\mathbf{3.47 \pm 2.82}$ & $\mathbf{0.27 \pm 0.27}$ \\
\midrule
MSVQ\_VAE & MSVQEnc & MSVQDec & $4.42 \pm 3.48$ & $4.02 \pm 3.21$ & $0.34 \pm 0.27$ \\
\bottomrule
\end{tabularx}
\renewcommand{\arraystretch}{1}
\end{table*}

Overall, the trajectory-based evaluation suggests that incorporating a GP prior improves temporal coherence relative to a plain VAE, leading to lower RMSE, ATE, and RPE on average. This effect is most pronounced for the PoseNet-based trajectory estimation, while the gains for classical visual odometry are smaller and more variable, indicating that GP regularization primarily benefits the preservation of temporally consistent motion cues rather than feature-based geometric matching alone.

\subsection{Latent Space}\label{subsec:latent}

\begin{figure*}[!tbp]
    \centering
    \includegraphics[width=\textwidth,height=0.82\textheight,keepaspectratio]{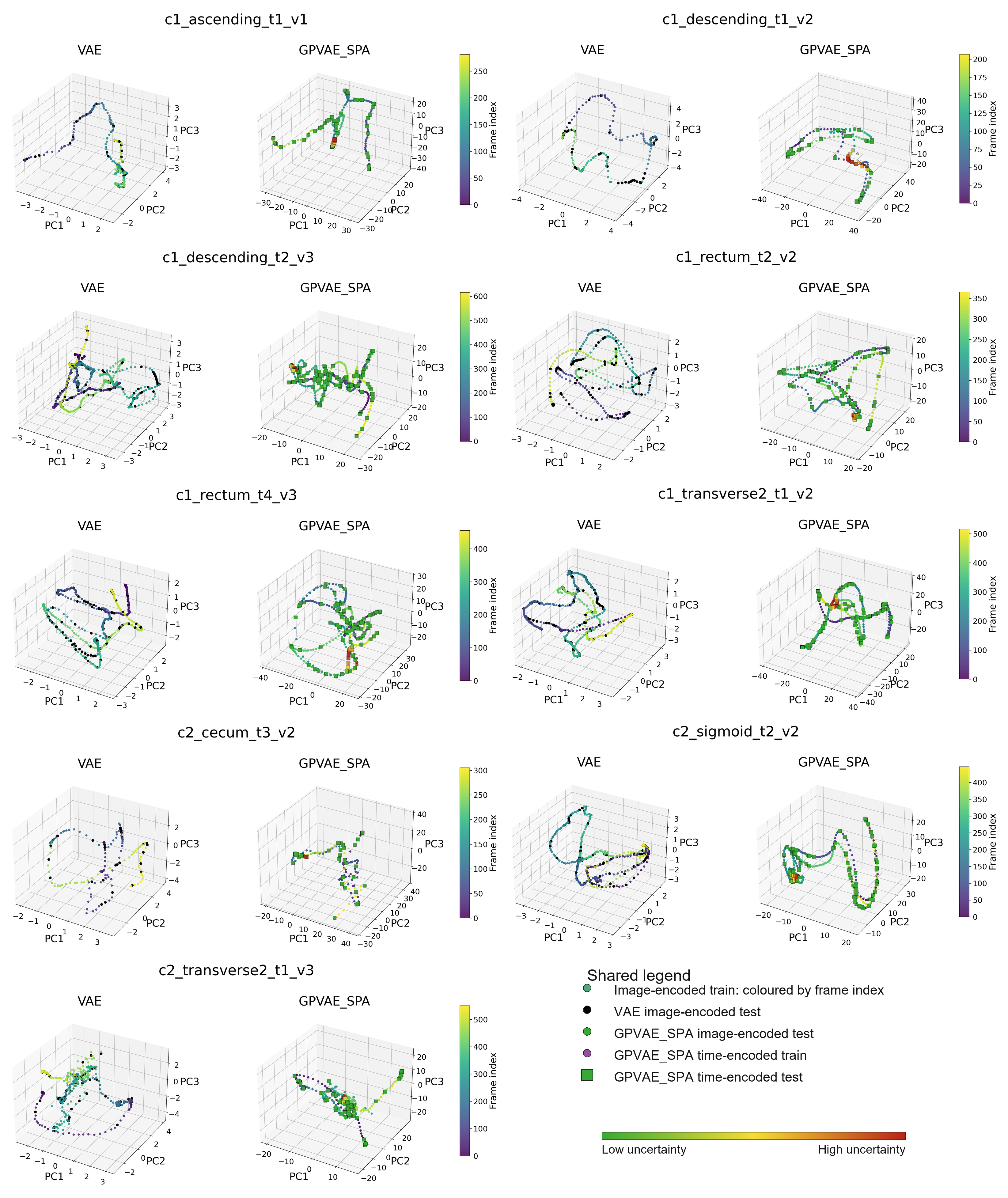}
    \caption{Three-dimensional PCA projections of image- and time-encoded latent trajectories for the VAE and GPVAE\_SPA models. Training points are colored by frame index, while held-out test points are colored by relative GP posterior uncertainty within each dataset, from low uncertainty in green to high uncertainty in red.}
    \label{fig:latent_tf20_gap10_spa}
\end{figure*}

Figure~\ref{fig:latent_tf20_gap10_spa} shows the latent trajectories obtained for all datasets when 20\% of the timestamps were held out from GP training. In addition, a contiguous gap of 10 consecutive frames was removed from each dataset to simulate realistic capsule endoscopy behavior, where bursts of missing frames can occur. All removed frames, including both the randomly held-out timestamps and the gap frames, were assigned to the test set.

For each dataset, the left panel shows the VAE latent trajectory obtained by encoding the image frames. The right panel shows the corresponding GPVAE\_SPA latent space and compares two representations: latent points obtained by encoding the frames with the image encoder, and latent points predicted from the timestamps alone using the GP prior. Training frames are colored by frame index. Held-out image-encoded test frames are shown as circular markers, while held-out time-encoded test frames in the GPVAE\_SPA panels are shown as square markers. The square markers are plotted larger than the training markers and on top of the training trajectory to make the test predictions visible. Both models used EndoVAEEnc and EndoVAEDec.

The main feature to inspect is the agreement between the image-encoded and time-encoded latent trajectories in the GPVAE\_SPA panels. When the square markers lie close to the image-encoded trajectory, the GP prior predicts a latent state from time alone that is consistent with the visual content of the corresponding frame. This illustrates the role of the temporal GP prior in GPVAE\_SPA: unlike a standard VAE, where the latent representation is obtained only through the image encoder, GPVAE\_SPA provides a second route into the latent space. A timestamp can be mapped to a latent distribution, which can then be decoded into an image.

The time-based latent prediction is also accompanied by the Gaussian process posterior uncertainty. We aggregate this uncertainty by computing the average GP posterior standard deviation across all latent coordinates, resulting in a single scalar uncertainty value per frame. For the held-out test frames, this scalar is encoded using a green--yellow--red color scale, where green indicates lower uncertainty and red indicates higher uncertainty. The color scale is normalized separately for each dataset using the uncertainty values of that dataset's held-out points. Therefore, colors should be interpreted within each panel only: red denotes relatively higher uncertainty for that dataset, while green denotes relatively lower uncertainty. The color scale is not shared in absolute value across datasets, so uncertainty colors should not be compared quantitatively between different datasets.

The uncertainty pattern provides an important diagnostic. Randomly held-out test frames are often surrounded by nearby training timestamps, and many of them therefore have relatively low uncertainty. This is expected: although their images were not used as GP training targets, their timestamps lie close to observed frames, so the GP prior can interpolate with relatively high confidence. In contrast, the contiguous gap creates a region of the temporal domain with no observed training frames. The GP uncertainty therefore tends to increase within the gap, especially for points farther away from the observed frames at the gap boundaries. This behavior is desirable, because it indicates that the model is less confident when it must infer latent states farther from available temporal support.

Thus, Figure~\ref{fig:latent_tf20_gap10_spa} should be read as both a trajectory-alignment plot and an uncertainty diagnostic. Good behavior corresponds to time-encoded test points that follow the image-encoded latent trajectory, while also showing increased uncertainty in genuinely unsupported regions such as the centre of a missing-frame gap. These uncertainty estimates can be interpreted as confidence measures for reconstructions generated from time alone, and may be useful for weighting predictions, flagging uncertain reconstructions, or deciding when image-based information is required.

\subsection{Effect of Latent Dimensionality}

To evaluate the influence of the latent dimensionality, we trained each model using latent dimensions
$D_z \in \{8,16,32,64,128,256\}$ and compared the reconstruction error on the training images.
The metric used is the root mean squared error (RMSE) between reconstructed and target training images,
reported in denormalized image-intensity space.

\begin{figure*}[!tbp]
    \centering
    \includegraphics[width=\textwidth,height=0.82\textheight,keepaspectratio]{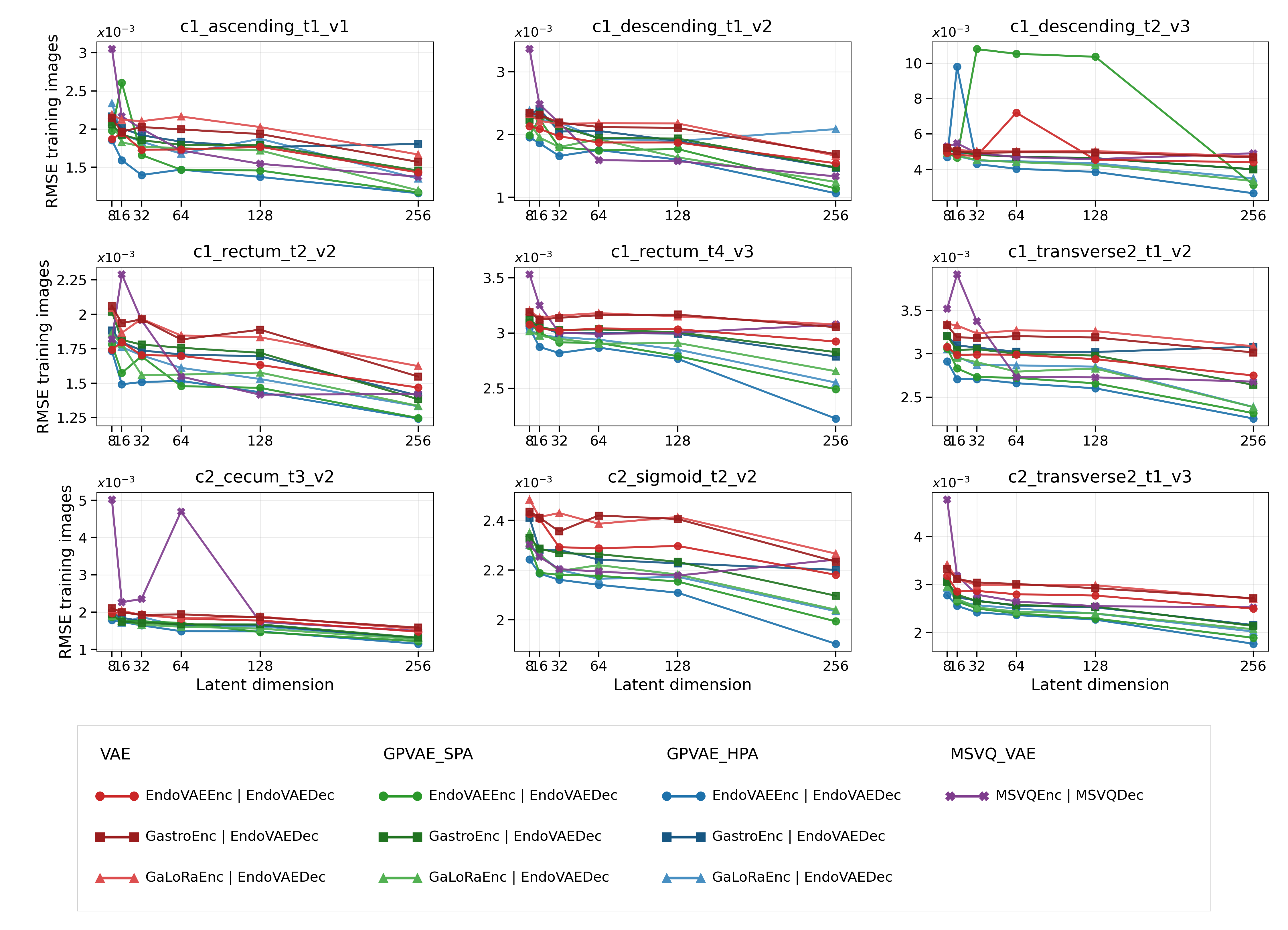}
    \caption{
    Training-image RMSE as a function of latent dimensionality for each dataset.
    Each subplot corresponds to one dataset. Colors indicate the model family, while markers indicate the encoder type.
    Lower values indicate better reconstruction quality. Increasing the latent space dimensionality results in lower RMSE.
    }
    \label{fig:rmse_latent_dim}
\end{figure*}

Overall, increasing the latent dimensionality tends to reduce the reconstruction error, although the improvement is not strictly monotonic for every dataset and model. The largest gains are generally observed when moving from very small latent spaces, such as 8D or 16D, to intermediate or high-dimensional latent spaces. The 256D setting often yields the lowest average RMSE, particularly for the GPVAE variants.

Across models, the GPVAE-based methods generally achieve lower RMSE than the plain VAE, suggesting that the structured latent prior remains beneficial even when reconstruction quality is evaluated directly in image space. The MSVQ-VAE results are more variable and should be interpreted with care. In the current implementation, MSVQ-VAE is not fully independent of the latent-dimensionality setting, because the global latent dimension is also used as the size of the continuous bottleneck before vector quantization. As a result, changing $D_z$ also changes the learned projection layers in the MSVQ-VAE model.

We also measured the average time per epoch for each model and latent dimensionality. This gives an indication of the computational cost associated with increasing the latent space size.

\begin{figure*}[!tbp]
    \centering
    \includegraphics[width=\textwidth,height=0.82\textheight,keepaspectratio]{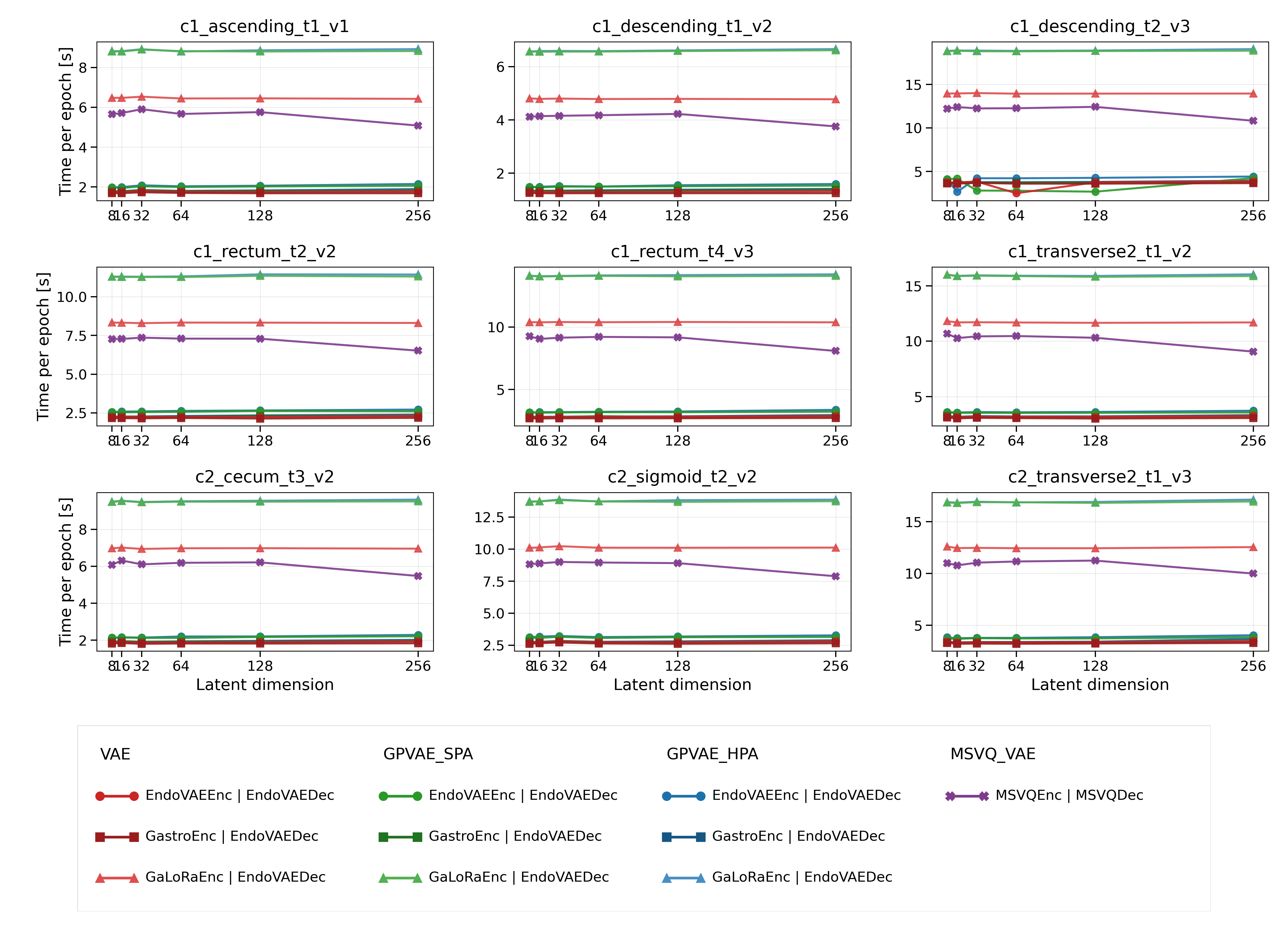}
    \caption{
    Time per epoch as a function of latent dimensionality for each dataset.
    Each subplot corresponds to one dataset. Colors indicate the model family, while markers indicate the encoder type. There is no effect of the latent space dimensionality on training time.
    }
    \label{fig:time_latent_dim}
\end{figure*}

The time-per-epoch results show that computational cost is driven more strongly by the model architecture and encoder choice than by the latent dimensionality itself. The plain VAE is consistently the fastest model family, while the GPVAE variants are slower due to the additional latent prior structure. MSVQ-VAE is generally the slowest, reflecting the cost of the multiscale encoder-decoder and vector-quantization bottleneck.

Importantly, increasing the latent dimensionality from 8D to 256D does not cause a strong systematic increase in time per epoch for most models. This suggests that, in these experiments, the dominant computational cost lies in the convolutional/encoder-decoder backbone and image-space losses rather than in the dimensionality of the latent vector. Therefore, larger latent spaces can improve reconstruction quality with only limited additional per-epoch cost.

\section{Discussion}\label{sec:discussion}

An important advantage of the GP-based formulation is that it provides uncertainty estimates over the latent trajectory and, indirectly, over the reconstructed frames. This uncertainty can serve as an intrinsic indicator of reconstruction reliability, for example by highlighting frames that are weakly supported by neighboring observations or by identifying interpolated frames for which the model is less confident. In this way, uncertainty can help distinguish between reconstructions that are plausible and well supported by the temporal context and those that are more speculative. Such information could be valuable in downstream applications, for instance by downweighting unreliable frames before pose estimation or $3$D reconstruction, or by prioritizing uncertain frames for human review in clinically sensitive settings. Although we do not yet exploit this uncertainty explicitly in the present work, it represents an important advantage of the proposed probabilistic framework and a promising direction for future research.

A main limitation of this work is that the experimental validation is performed primarily on synthetic endoscopic video with controlled ground truth. This enables rigorous quantitative evaluation of reconstruction quality, missing-frame inference, uncertainty, and trajectory preservation, but it does not fully capture the variability and complexity of real clinical endoscopy. In practice, real procedures exhibit stronger domain shifts, non-rigid tissue deformation, illumination changes, specularities, occlusions, blur, and procedure-dependent motion patterns that may affect both the learned latent dynamics and the reliability of downstream pose estimation. As a result, although the proposed framework shows promising performance on C3VDv2, further validation on real clinical video data will be necessary to assess the extent to which these gains transfer to practical endoscopic applications.

In this work, the Gaussian-process prior is indexed by time, implemented here as the frame number. However, the proposed framework is not restricted to temporal inputs alone. In principle, any auxiliary variable that provides meaningful structure over the sequence could be used to define the GP input space. For example, in wireless capsule endoscopy, external magnetic localization systems can provide position and orientation estimates of the capsule, which could serve as physically grounded covariates for the latent process \citep{Xu2022AEndoscopy, Kanaan2025EnablingLocalization}. Likewise, inertial measurements or other onboard sensing modalities could be incorporated when available, as sensor fusion has also been explored for capsule localization \citep{Vedaei2021ASensor}. Beyond physical sensing, clinically meaningful sequence-level information, such as anatomical segment labels, procedural phase, or inspection context, could also be used to structure the latent trajectory. Such auxiliary inputs may help the model capture temporal dependencies more faithfully and could potentially improve downstream tasks such as camera path reconstruction by providing a latent prior that is better aligned with the underlying acquisition process.

\section{Conclusion}\label{sec:conclusion}

We presented a Gaussian Process Prior Variational Autoencoder framework tailored to the challenges of endoscopic video restoration. By replacing the independent latent prior of a standard VAE with a temporal Gaussian process prior, the model transforms the latent representation from a collection of independent points into a continuous probabilistic trajectory over time. This design enables two complementary capabilities: reconstructing frames from their visual content while exploiting temporal context, and inferring plausible latent states, and thus reconstructed frames, from timestamps alone when image observations are absent or heavily corrupted.

Experiments on the C3VDv2 colonoscopy dataset demonstrate that the GP prior consistently improves reconstruction quality over the plain VAE baseline, with GPVAE-HPA using the EndoVAE encoder achieving the strongest pixel-level performance. Across matched encoder--decoder groups, the best GPVAE approximation reduced image reconstruction RMSE by 21.9\% on average, with a maximum reduction of 26.1\%, relative to the corresponding VAE baselines. The time-encoded reconstruction results confirm that the learned latent trajectories capture meaningful temporal structure rather than merely acting as a regulariser during training. Camera path evaluation using both classical visual odometry and a pretrained PoseNet shows that GPVAE variants better preserve the temporal motion cues required for downstream geometric tasks, with the most pronounced gains observed under the learned PoseNet estimator. Across both trajectory estimators, the best GPVAE approximation reduced trajectory RMSE by 12.7\% on average relative to the corresponding VAE baselines. The GP posterior uncertainty provides an additional diagnostic that reflects the degree of temporal support available for each reconstructed frame, with uncertainty appropriately increasing within contiguous missing-frame gaps and decreasing for frames surrounded by nearby observations.

Several directions remain open for future work. Validation on real clinical endoscopic video is a necessary next step to assess the extent to which the gains observed on synthetic data transfer to the variability and complexity of live procedures. The GP input space need not be restricted to frame indices: physically grounded covariates such as magnetic capsule localisation signals, inertial measurements, or anatomical segment labels could provide richer temporal structure and better align the latent prior with the underlying acquisition process. Finally, the uncertainty estimates produced by the GP posterior have not yet been exploited explicitly in downstream tasks. Integrating them as confidence weights for pose estimation, 3D reconstruction, or clinical review represents a promising and practically meaningful direction.


\section*{Declaration of Competing Interest}
\noindent The authors declare that they have no known competing financial interests or personal relationships that could have appeared to influence the work reported in this paper. 

\section*{Funding}
\noindent Ivan De Boi is funded by the FWO postdoc fellowship 1217125N.

\appendix

\section{Implementation Details of the Objective}
\label{app:objective}

\subsection{Reconstruction Objective}
\label{app:reconstruction_objective}

In \citep{Shi2025Neighbour-DrivenModelling}, the reconstruction term is written as the negative log-likelihood of the decoder distribution, with the likelihood chosen according to the observation type: a Bernoulli likelihood for binary image data and a Gaussian likelihood for continuous-valued observations.

In our implementation, the decoder outputs an RGB reconstruction directly, and we optimize a deterministic reconstruction loss rather than an explicit probabilistic observation likelihood. Instead, we train the VAE and GPVAE variants with a weighted reconstruction objective in image space:
\begin{equation}
\mathcal{L}_{\mathrm{total}}
=
\lambda_{\mathrm{MSE}}\mathcal{L}_{\mathrm{MSE}}
+
\lambda_{\mathrm{LPIPS}}\mathcal{L}_{\mathrm{LPIPS}}
+
\lambda_{\mathrm{KL}}\beta_t\mathcal{L}_{\mathrm{KL}},
\end{equation}
where $\lambda_{\mathrm{MSE}}=4.0$, $\lambda_{\mathrm{LPIPS}}=2.0$, and $\lambda_{\mathrm{KL}}=1.0$. The weights were chosen empirically. The KL warm-up factor $\beta_t$ is annealed during training. The KL term remains a Gaussian latent-space regularizer. For the standard VAE it is computed against an independent standard normal prior, while for the GPVAE variants it is replaced by the corresponding neighbor-based GP prior term.

The pixel-wise term is the mean-squared error between the target frame $\mathbf{y}$ and reconstruction $\mathbf{y}'$:
\begin{equation}
\mathcal{L}_{\mathrm{MSE}}
=
\frac{1}{|\Omega|}
\sum_{p \in \Omega}
\left\|
\mathbf{y}_p - \mathbf{y}'_p
\right\|_2^2 ,
\end{equation}
where $\Omega$ is the set of pixels included in the loss.

The perceptual term is
\begin{equation}
\mathcal{L}_{\mathrm{LPIPS}}
=
\mathrm{LPIPS}(\mathbf{y},\mathbf{y}'),
\end{equation}
computed with a fixed VGG-based LPIPS network. LPIPS is evaluated after resizing images to $256 \times 256$ for computational efficiency, whereas the MSE term is computed at the full training resolution.

The KL term $\mathcal{L}_{\mathrm{KL}}$ varies between the models and is explained in the next section.

\subsection{KL Terms}
\label{app:kl_terms}

For the standard VAE, the encoder outputs a diagonal Gaussian approximate posterior,
\begin{equation}
q_\phi(z_i \mid y_i)
=
\mathcal{N}\left(
z_i;
\mu_\phi(y_i),
\operatorname{diag}(\sigma^2_\phi(y_i))
\right),
\end{equation}
and the prior is the standard normal distribution,
\begin{equation}
p(z_i)=\mathcal{N}(0,I).
\end{equation}
The KL term is therefore computed independently for each frame and averaged over the mini-batch:
\begin{equation}
\mathcal{L}_{\mathrm{KL}}^{\mathrm{VAE}}
=
\frac{1}{B}
\sum_{i=1}^{B}
\frac{1}{2}
\sum_{d=1}^{D}
\left(
\mu_{i,d}^2
+
\sigma_{i,d}^2
-
1
-
\log \sigma_{i,d}^2
\right),
\end{equation}
where $B$ is the batch size and $D$ is the latent dimensionality. In the implementation, the log-variance is clamped to $[-10,10]$ for numerical stability.

For GPVAE-HPA, the KL term is computed on local neighborhood blocks. For each batch index $i$, we construct a local set
\begin{equation}
\mathcal{N}_i = \{i\} \cup n(i),
\end{equation}
where $n(i)$ contains the $H$ nearest neighbors in the GP input space. Thus, the local block size in our implementation is $K=H+1$. For each latent dimension $d$, the approximate posterior over this local block is
\begin{equation}
q_\phi(z_{\mathcal{N}_i}^{(d)} \mid Y_{\mathcal{N}_i})
=
\mathcal{N}
\left(
\mu_{\mathcal{N}_i}^{(d)},
\operatorname{diag}\left((\sigma_{\mathcal{N}_i}^{(d)})^2\right)
\right),
\end{equation}
while the GP prior over the same block is
\begin{equation}
p_\theta(z_{\mathcal{N}_i}^{(d)} \mid X_{\mathcal{N}_i})
=
\mathcal{N}
\left(
0,
K_{\theta}(X_{\mathcal{N}_i},X_{\mathcal{N}_i}) + \sigma_n^2 I
\right).
\end{equation}
The local HPA KL is computed in closed form:
\begin{equation}
\begin{aligned}
\mathrm{KL}_{i,d}^{\mathrm{HPA}}
=
\frac{1}{2}
\Big[
&\operatorname{tr}\!\left(K_i^{-1} S_{i,d}\right)
+
\mu_{i,d}^{\top} K_i^{-1} \mu_{i,d}
\\
&-
\log \left|S_{i,d}\right|
+
\log \left|K_i\right|
-
K
\Big].
\end{aligned}
\end{equation}
where $K_i$ is the local GP covariance matrix and $S_{i,d}$ is the diagonal posterior covariance for latent dimension $d$ on the local block. In the code, this term is averaged over batch elements and latent dimensions, and additionally divided by the local block size:
\begin{equation}
\mathcal{L}_{\mathrm{KL}}^{\mathrm{HPA}}
=
\frac{1}{B D K}
\sum_{i=1}^{B}
\sum_{d=1}^{D}
\mathrm{KL}_{i,d}^{\mathrm{HPA}}.
\end{equation}

For GPVAE-SPA, the GP prior is approximated through local conditional distributions. For each frame $i$, let $n(i)$ denote its nearest-neighbor set. The GP conditional prior for latent dimension $d$ is
\begin{equation}
p_\theta(z_i^{(d)} \mid z_{n(i)}^{(d)})
=
\mathcal{N}
\left(
\alpha_i^\top z_{n(i)}^{(d)},
v_i
\right),
\end{equation}
where
\begin{equation}
\alpha_i = K_{n(i)n(i)}^{-1}K_{n(i)i},
\end{equation}
and
\begin{equation}
v_i =
K_{ii} - K_{i n(i)}K_{n(i)n(i)}^{-1}K_{n(i)i}.
\end{equation}
In the implementation, the conditional mean is approximated using the detached encoder means of the neighbors:
\begin{equation}
m_{i,d}
=
\alpha_i^\top \mu_{n(i),d}.
\end{equation}
The KL between the current approximate posterior and this local conditional prior is
\begin{equation}
\mathrm{KL}_{i,d}^{\mathrm{SPA}}
=
\frac{1}{2}
\left[
\log\frac{v_i}{\sigma_{i,d}^2}
+
\frac{\sigma_{i,d}^2 + (\mu_{i,d}-m_{i,d})^2}{v_i}
-
1
\right].
\end{equation}
The implementation averages this term over the batch and latent dimensions, and rescales it by $N/B$:
\begin{equation}
\mathcal{L}_{\mathrm{KL}}^{\mathrm{SPA}}
=
\frac{N}{B}
\cdot
\frac{1}{BD}
\sum_{i=1}^{B}
\sum_{d=1}^{D}
\mathrm{KL}_{i,d}^{\mathrm{SPA}}.
\end{equation}

Our implementation follows the neighbor-driven HPA idea of \citep{Shi2025Neighbour-DrivenModelling}, but differs in several practical details. First, our notation for the neighborhood size differs slightly: we use $H$ for the number of neighbors excluding the current frame. Second, our HPA implementation rescales the local KL by averaging over the mini-batch and latent dimensions and dividing by the local block size. This keeps the numerical scale of the KL term stable across batch size, latent dimensionality, and neighborhood size, but means that the loss is not exactly the same mini-batch ELBO estimator as in the original formulation. 
Finally, for efficiency, the encoder means and log-variances for all frames are precomputed once per epoch and treated as fixed when evaluating the local HPA KL term.

Our SPA implementation follows the neighbor-driven conditional-prior idea of \citep{Shi2025Neighbour-DrivenModelling}, but uses several practical approximations. As for HPA, $H$ denotes the number of neighbors excluding the current frame. For each batch element $i$, the conditional prior is computed for the current frame given its neighbor set $n(i)$. These neighbors are selected as nearest neighbors in the full temporal input space, rather than being restricted to preceding points as in a strictly ordered Vecchia approximation. This choice is appropriate for our interpolation setting, where frames on both sides of a missing observation may provide temporal context.

For efficiency, the encoder means for all frames are precomputed once per epoch and treated as fixed when constructing the SPA conditional mean. The current batch posterior parameters remain part of the computational graph, so the SPA KL still regularizes the encoder output for the current batch. The resulting KL is averaged over the batch and latent dimensions and rescaled by $N/B$ to approximate a full-sequence contribution.

\section{Path Reconstruction} \label{app:pathrecon}

\subsection{Visual Odometry}

Classical visual odometry estimates camera motion directly from image correspondences between successive or nearby frames. Given two frames $I_i$ and $I_j$, salient keypoints are first detected and matched, for which we rely on ORB features. C3VDv2 provides images of checkerboards, which we used to calibrate the camera. Using the known camera intrinsics encoded in the calibration matrix $\mathbf{K}$, the matched image points are used to estimate the essential matrix $\mathbf{E}$, which captures the epipolar geometry between the two calibrated views. For corresponding normalized image points $\mathbf{x}_i$ and $\mathbf{x}_j$, this relation satisfies
\begin{equation}
\mathbf{x}_j^\top \mathbf{E}\,\mathbf{x}_i = 0.
\end{equation}
Decomposition of $\mathbf{E}$ then yields the relative rotation and translation direction between the two camera views, which together define the relative rigid transform. More details can be found in \citep{Hartley2004MultipleVision}.

\subsection{Root Mean Squared Error}

The root mean squared error (RMSE) measures the average squared positional deviation between the aligned estimated trajectory and the ground-truth trajectory.:
\begin{equation}
\mathrm{RMSE}
=
\sqrt{
\frac{1}{N}
\sum_{t=1}^{N}
\left\|
\bar{\mathbf{p}}_t - \mathbf{p}^{\mathrm{gt}}_t
\right\|_2^2
},
\end{equation}
where $\bar{\mathbf{p}}_t$ is the aligned estimated camera position at frame $t$, $\mathbf{p}^{\mathrm{gt}}_t$ is the corresponding ground-truth position, and $N$ is the number of shared frames. Lower RMSE indicates better overall agreement with the ground-truth path.

\subsection{Absolute Trajectory Error}

The absolute trajectory error (ATE) measures the mean Euclidean distance between aligned estimated positions and the ground truth:
\begin{equation}
\mathrm{ATE}
=
\frac{1}{N}
\sum_{t=1}^{N}
\left\|
\bar{\mathbf{p}}_t - \mathbf{p}^{\mathrm{gt}}_t
\right\|_2.
\end{equation}
While RMSE penalizes larger errors more strongly, ATE provides a more direct measure of the average absolute positional deviation along the trajectory.

\subsection{Relative Pose Error}

The relative pose error (RPE) evaluates how well local frame-to-frame motion is preserved. Defining the frame-to-frame translation increments as
\begin{equation}
\Delta \bar{\mathbf{p}}_t = \bar{\mathbf{p}}_t - \bar{\mathbf{p}}_{t-1},
\qquad
\Delta \mathbf{p}^{\mathrm{gt}}_t = \mathbf{p}^{\mathrm{gt}}_t - \mathbf{p}^{\mathrm{gt}}_{t-1},
\end{equation}
the RPE is computed as
\begin{equation}
\mathrm{RPE}
=
\sqrt{
\frac{1}{N-1}
\sum_{t=2}^{N}
\left\|
\Delta \bar{\mathbf{p}}_t - \Delta \mathbf{p}^{\mathrm{gt}}_t
\right\|_2^2
}.
\end{equation}
This metric is sensitive to local drift and indicates how accurately the estimated trajectory reproduces the relative motion between consecutive frames.

\bibliographystyle{plainnat}

\bibliography{references}



\end{document}